\documentclass[10pt,final,journal]{IEEEtran} 

\pdfoutput=1
\newcommand{\xz}{\textcolor{red}}
\newcommand{\xzb}{\textcolor{blue}}
\newcommand{\xzg}{\textcolor{green}}

\usepackage{lineno,hyperref}
\usepackage{moreverb}
\usepackage{fancyhdr}
\usepackage{calc}
\usepackage{float}
\usepackage{subfigure}
\usepackage{color}
\usepackage[table]{xcolor}
\usepackage{amsmath}
\usepackage{arydshln}

\usepackage{graphicx}
\usepackage{epstopdf}

\usepackage{bm}
\usepackage{setspace}
\usepackage{multirow}

\usepackage[numbers,sort&compress]{natbib} 
\usepackage{rotating}

\ifCLASSINFOpdf
\else
\fi

\begin{document}
	%
	\title{Multi-focus Image Fusion: A Benchmark}

	%
	%
	%
	
	\author{Xingchen~Zhang \IEEEmembership{}
			\thanks{X.~Zhang is with the Department of Electrical and Electronic Engineering,
				Imperial College London, London, United Kingdom. e-mail: xingchen.zhang@imperial.ac.uk}
	}

	%
	%
	
	\markboth{Journal of \LaTeX\ Class Files,~Vol.~XX, No.~XX, XX~2020}%
	{Shell \MakeLowercase{\textit{et al.}}: Bare Demo of IEEEtran.cls for IEEE Journals}
	%
	
	
	
	\maketitle
	
	\begin{abstract}
 Multi-focus image fusion (MFIF) has attracted considerable interests due to its numerous applications.~While much progress has been made in recent years with efforts on developing various MFIF algorithms, some issues significantly hinder the fair and comprehensive performance comparison of MFIF methods, such as the lack of large-scale test set and the random choices of objective evaluation metrics in the literature.~To solve these issues, this paper presents a multi-focus image fusion benchmark (MFIFB) which consists a test set of 105 image pairs, a code library of 30 MFIF algorithms, and 20 evaluation metrics.~MFIFB is the first benchmark in the field of MFIF and provides the community a platform to compare MFIF algorithms fairly and comprehensively.~Extensive experiments have been conducted using the proposed MFIFB to understand the performance of these algorithms.~By analyzing the experimental results,  effective MFIF algorithms are identified.~More importantly, some observations on the status of the MFIF field are given, which can help to understand this field better.
	\end{abstract}


	%
	\IEEEpeerreviewmaketitle
	
\section{Introduction}
\begin{figure}
	\centering
	\includegraphics[width=6.5cm]{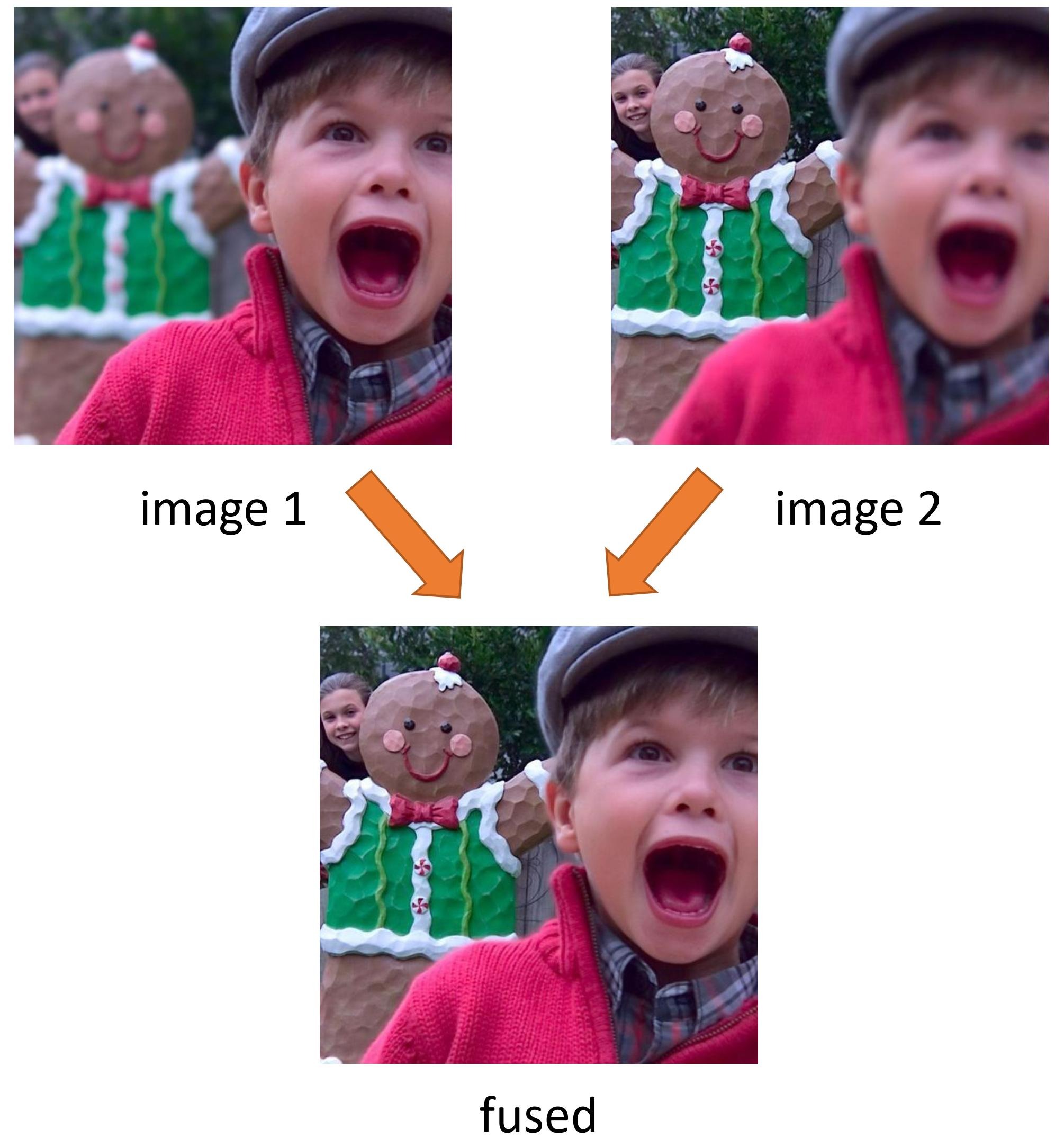}
	\caption{The benefit of multi-focus image fusion.~In image 1, the background is not clear while in image 2 the foreground is not clear.~After fusion, both the background and foreground in the fused image are clear.~The fused image is produced by CBF \cite{kumar2015image}.}
	\label{fig:example-fusion}
\end{figure}

Clear images are desirable in computer vision applications.~However, it is difficult to have all objects in focus in an image since most imaging systems have a limited depth-of-field (DOF).~To be more specific, scene contents within the DOF remain clear while objects outside that area appear as blurred.~Multi-focus image fusion (MFIF) aims to combine multiple images with different focused areas into a single image with everywhere in focus, as shown in Fig.~\ref{fig:example-fusion}.

MFIF has attracted considerable interests recently and various MFIF algorithms have been proposed, which can be generally divided into spatial domain-based methods and transform domain-based methods.~Spatial domain-based methods operate directly in spatial domain and can be roughly divided into three categories: pixel-based \cite{liu2015multi}, block-based \cite{de2013multi} and region-based \cite{li2006region}.~In contrast, transform domain-based methods firstly transform images into another domain and then perform fusion in that transformed domain.~The fused image is then obtained via the inverse transformation.~The representative transform domain-based methods are sparse representation (SR) methods \cite{zhang2018robust, zhang2018sparse} and multi-scale methods \cite{amin2018multi, kou2018multi}.

In recent years, with the development of deep learning, researchers have begun to solve the MFIF problem with deep learning techniques.~Both supervised \cite{zhai2020multi, zhang2020ifcnn, wang2020novel, xu2020deep} 
and unsupervised MFIF algorithms \cite{yan2018unsupervised, ma2019sesf, mustafa2019dense, jung2020unsupervised} have been proposed.~To be more specific, various deep learning models and methods have been employed, such as CNN \cite{li2018convolutional, du2017image}, GAN \cite{guo2019fusegan}
and ensemble learning \cite{amin-naji2019ensemble}.

\begin{table*}
	\begin{center}
		\caption{Some MFIF algorithms published in top journals and conferences.~The number of tested image pairs, the number of compared algorithms, the number of utilized evaluation metrics are also given.~The details of the proposed MFIFB are also shown.}
		\label{table:published-some}
		\scriptsize
		\renewcommand\arraystretch{1.5}
		\begin{tabular}{llllll}		
			\hline
			Reference      & Year   & Journal     & Image pairs  & Algorithms  & Metrics \\ \hline
			GFF \cite{li2013image}  &  2013   &   IEEE Transactions on Image Processing                 & 10 & 7  & 5 (MI, $Q_Y$, $Q_C$, $Q_G$, $Q_P$)\\
			RP\textunderscore SR \cite{liu2015general}   &  2015 &  Information Fusion  &10  & 8  & 5 (SD, EN, $Q_G$, $Q_P$, $Q_W$)\\
			MST\textunderscore SR \cite{liu2015general}  & 2015  & Information Fusion   &10  & 8  & 5 (SD, EN, $Q_G$, $Q_P$, $Q_W$)\\	 
			NSCT\textunderscore SR \cite{liu2015general}   &2015   & Information Fusion &10  & 8  & 5 (SD, EN, $Q_G$, $Q_P$, $Q_W$) \\
			QB \cite{bai2015quadtree} & 2015& Information Fusion  & 6  & N/A  & 3 ($Q_{GM}$, $Q^{AB/F}$, NMI)\\
			DSIFT \cite{liu2015multi} & 2015  & Information Fusion  &  12 & 9  & 6 (NMI, $Q_{NCIE}$, $Q_G$, PC, $Q_Y$, $Q_{CB}$) \\	
			MRSR \cite{zhang2016robustmulti} & 2016 & IEEE Transactions on Image Processing &7  & 9 & 4 (MI, $Q_G$, ZNCC\textunderscore PC, $Q_{PC}$)   \\
			CNN \cite{liu2017multi}  & 2017   &  Information Fusion     &40  &  6 & 4 (NMI, $Q^{AB/F}$, $Q_Y$, $Q_{CB}$) \\	
			BFMF \cite{zhang2017boundary} &   2017 &    Information Fusion & 18 & 6 & 4 (NMI, PC, $Q_{MSSI}$, $Q_C$) \\
			\cite{zhang2018sparse} & 2018 & Information Fusion  & 10 &9 & 5 (MI, $Q_G$, $Q_S$, $Q_{ZP}$, $Q_{PC}$ ) \\
			p-CNN \cite{tang2018pixel}  & 2018  &  Information Science &   12 & 5 &4 (NMI, $Q_{PC}$, $Q_W$, $Q_{CB}$) \\
			CAB \cite{farid2019multi}  & 2019  & Information Fusion    & 34 & 15 &  5 ($Q^{AB/F}$, NMI, FMI, $Q_Y$, $Q_{NCIE}$) \\
			mf-CRF \cite{bouzos2019conditional} & 2019 & IEEE Transactions on Image Processing &52 &11 & 4 (MI, $Q_G$, $Q_Y$, $Q_{CB}$)\\
			DIF-Net \cite{jung2020unsupervised}  & 2020 & IEEE Transactions on Image Processing &20 & 9& 7 (MI, FMI, $Q_X$, $Q_{SCD}, $ $Q_H$, $Q_P$, $Q_M$) \\
			DRPL \cite{li2020drpl}   & 2020   &  IEEE Transactions on Image Processing    & 20 & 7  & 5 (MI, $Q^{AB/F}$, AG, VIF, EI) \\ 	
			IFCNN \cite{zhang2020ifcnn}  & 2020 & Information Fusion& 20 & 4   & 5 (VIFF, ISSIM, NMI, SF, AG) \\
			FusionDN \cite{xu2020fusiondn}  & 2020  & AAAI & 10  & 5   &  4 (SD, EN, VIF, SCD)       \\
			PMGI \cite{zhang2020PMGI}  & 2020 & AAAI & 18 & 5  & 6 (SSIM, $Q^{AB/F}$, EN, FMI, SCD, CC) \\ \hline 
			\textbf{MFIFB}                      & 2020 &      & \textbf{105} & \textbf{30} &\textbf{20} (CE, EN, FMI, NMI, PSNR, $Q_{NCIE}$, TE, AG, EI, $Q^{AB/F}$, $Q_P$, \\ 
			                           &        &      &   &   & ~~~~~SD, SF, $Q_C$, $Q_W$, $Q_Y$, SSIM, $Q_{CB}$, $Q_{CV}$, VIF)                                \\   
 			\hline 
		\end{tabular}
	\end{center}
\end{table*}

However, current research on MFIF is suffering from several issues, which hinder the development of this field severely.~First, there is not a well-recognized MFIF benchmark which can be used to compare performance under the same standard.~Therefore, it is quite common that different images are utilized in experiments in the literature, which makes it difficult to fairly compare the performance of various algorithms.~Although the Lytro dataset \cite{nejati2015multi} is used frequently, many researchers only choose several image pairs from it in experiments, resulting in bias results.~This is very different from other image processing-related areas like visual object tracking where several benchmarks \cite{wu2015object, VOT_TPAMI} are available and every paper has to show results on some of them.~Second,
as the most widely used dataset, the Lytro dataset only consists of 20 pairs of multi-focus images which are not enough for large-scale comparison.~Also, Xu et al.~\cite{xu2020mffw} showed that the defocus spread effect (DSE) is not obvious in Lytro dataset thus popular methods perform very similar on it.~Third, although many evaluation metrics have been proposed to evaluate the image fusion algorithms, none of them is better than all other metrics.~As a result, researchers normally choose several metrics which support their methods in the literature.~This makes it not trivial to compare performances objectively.~Table \ref{table:published-some} lists some algorithms published in top journals (conferences) and the number of image pairs, compared algorithms, and evaluation metrics.~As can be seen, these works present results of different evaluation metrics on different number of image pairs, making it quite difficult to ensure a fair and comprehensive performance comparison.~Besides,  many researchers only choose several algorithms which may be outdated to compare with their own algorithms, making it more difficult to know the real performance of these algorithms.~More importantly, it is frequent that methods are compared with those which are not designed for this task \cite{deng2019deep}.~For example, the performance of a MFIF algorithm is compared with a method designed for visible-infrared image fusion.~Finally, although the source codes of some MFIF algorithms have been made publicly available, the usage of these codes are different.~For examples, different codes have different interfaces to read and write images, and may have various dependencies to install.~Therefore, it is inconvenient and time-consuming to conduct large scale performance evaluation.~It is thus desirable that results on public datasets are available and a consistent interface is available to integrate new algorithms conveniently for performance comparison.

To solve these issues, in this paper a multi-focus image fusion benchmark (MFIFB) is created, which includes 105 pairs of multi-focus images, 30 publicly available fusion algorithms, 20 evaluation metrics and an interface to facilitate the algorithm running and performance evaluation.~The main contributions of this paper lie in the following aspects:
\begin{itemize}	
	\item \textbf{Dataset}.~A test set containing 105 pairs of multi-focus images is created.~These image pairs cover a wide range of environments and conditions.~Therefore, the test set is able to test the generalization ability of fusion algorithms.
	
	\item \textbf{Code library}.~30 recent MFIF algorithms are collected and integrated into a code library, which can be easily utilized to run algorithms and compare performances.~An interface is designed to integrate new image fusion algorithms into MFIFB.~It is also convenient to compare performances using fused images produced by other algorithms with those available in MFIFB.

	\item \textbf{Comprehensive performance evaluation}.~20 evaluation metrics are implemented in MFIFB to comprehensively compare fusion performance.~This is much more than those utilized in the MFIF literature as shown in Table \ref{table:published-some}.~Extensive experiments have been conducted using MFIFB, and the comprehensive comparison of those algorithms are performed.
	
\end{itemize}

The rest of this paper is organized as follows.~Section \ref{sec:review} gives some background information about MFIF.~Then, the proposed multi-focus image fusion benchmark is introduced in detail in Section \ref{sec:benchmark}, followed by experiments and analysis in Section \ref{sec:experiment}.~Finally, Section \ref{sec:conclusion} concludes the paper.

\section{Multi-focus image fusion methods}
\label{sec:review}
\subsection{The background of multi-focus image fusion}
MFIF aims to produce an all-in-focus image by fusing multiple partially focused images of the same scene \cite{fu2020novel}.~Normally, MFIF is solved by combining focused regions with some fusion rules.~The key task in MFIF is thus the identification of focused and defocused area, which is normally formulated as a classification problem.

Various focus measurements (FM) were designed to classify whether a pixel is focused or defocused.~For example, Zhai et al.~\cite{zhai2020multi} used the energy of Laplacian to detect the focus level of source images.~Tang et al.~\cite{tang2018pixel} proposed a pixel-wise convolutional neural network (p-CNN) which was a learned FM that can recognize the focused and defocused pixels.

\begin{figure*}
	\centering
	\includegraphics[width=0.9\textwidth]{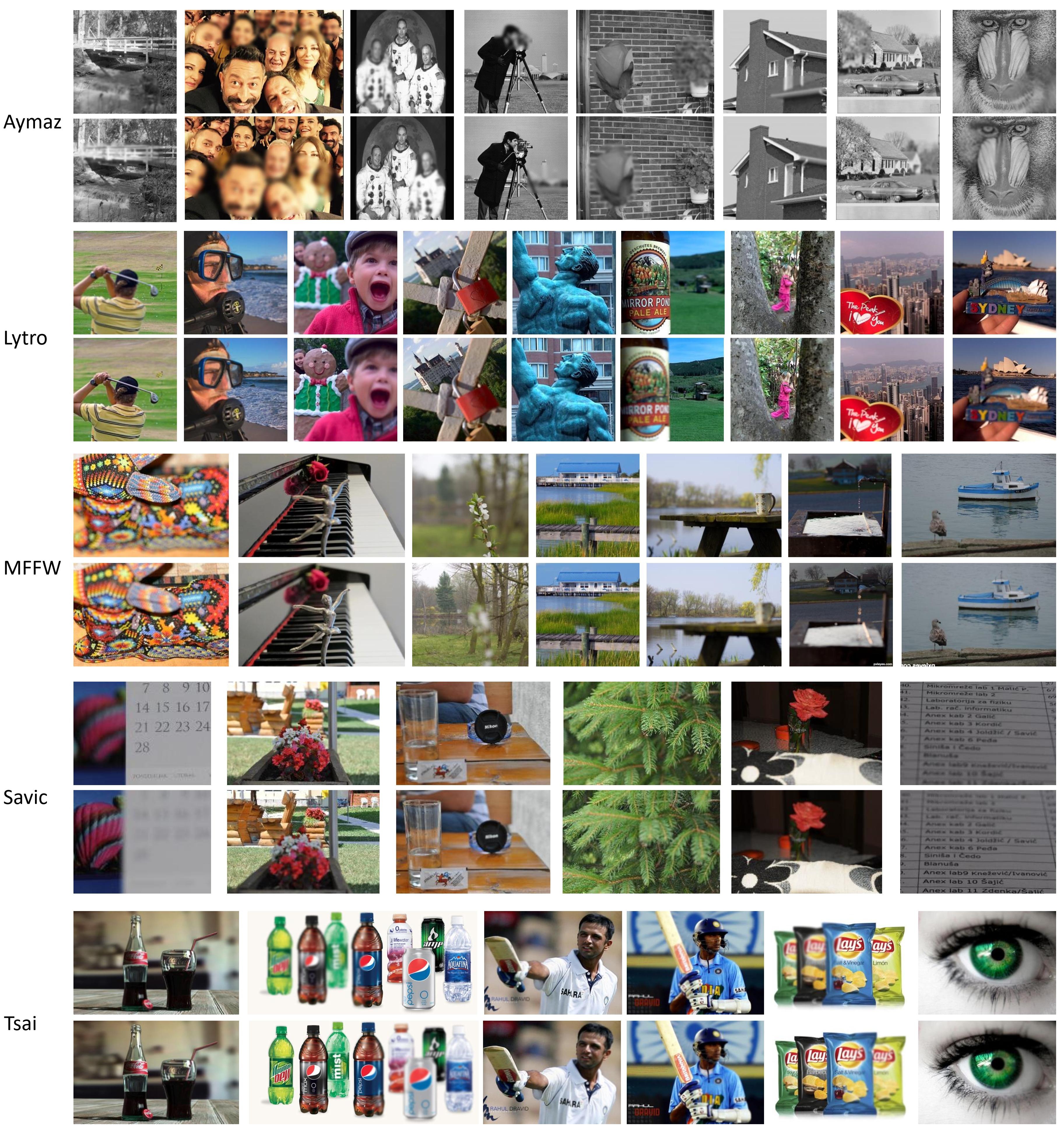}
	\caption{A part of dataset in MFIFB.}
	\label{fig:example}
\end{figure*}

\subsection{Conventional multi-focus image fusion methods}

Generally speaking, conventional MFIF algorithms can be divided into 
spatial domain-based methods and transform domain-based methods.~Spatial domain-based methods operate directly in spatial domain.~According to the adopted ways, these methods can be classified as pixel-based \cite{liu2015multi}, block-based \cite{de2013multi} and region-based \cite{li2006region}.~In pixel-based methods, the FM is applied at pixel-level to judge whether a pixel is focused or defocused.~In block-based methods, the source images are firstly divided to blocks with fixed size, and then the FM is applied to these patches to decide their blurring levels.~However, the performance of block-based methods heavily dependent on the division of blocks and may induce artifacts easily.~In region-based methods, the source images are firstly segmented into different regions using segmentation techniques, and then the blurring levels of these regions are calculated based on FM.

The transform domain-based methods normally consist of three steps.~First, the source images are transformed to another domain using some transformations, such as wavelet transform and sparse representation.~The source images can be represented using some coefficients in this way.~Second, the coefficients of source images are fused using designed fusion rules.~Finally, the fused image is obtained by applying inverse transformation to those fused coefficients.~Transform domain-based algorithms mainly contain sparse representation-based \cite{yang2010multi, liu2014simultaneous}, multi-scale-based \cite{wang2003multifocus, zhi2004wavelet}, subspace-based \cite{bavirisetti2017multi}, edge-preserving-based \cite{chen2017robust, zhao2017multisensor} and others.

\subsection{Deep learning-based methods}
In recent years, deep learning has been applied to MFIF and an increasing number of deep learning-based methods emerge every year.~Liu et al.~\cite{liu2017multi} proposed the first CNN-based method which utilized a CNN to learn a mapping from source images to the focus map.~Since then, more than 40 deep learning-based MFIF algorithms have been proposed.

The majority of deep learning-based MFIF algorithms are supervised algorithms, which need a large amount of training data with ground-truth to train.~For instance, Zhao et al.~\cite{zhao2018multi} developed a MFIF algorithm based on multi-level deeply supervised CNN (MLCNN).~Tang et al.~\cite{tang2018pixel} proposed a pixel-wise convolutional neural network (p-CNN) which was a learned FM that can recognize the focused and defocused pixels in source images.~Yang et al.~\cite{yang2019multilevel} proposed a multi-level features convolutional neural network (MLFCNN) architecture for MFIF.~Li et al.~\cite{li2020drpl} proposed the DRPL, which directly converts the whole image into a binary mask without any patch operation.~Zhang et al.~\cite{zhang2020ifcnn} proposed a general image fusion framework based on CNN (IFCNN).

In supervised learning-based methods, a large amount of labeled training data is needed, which is labor-intensive and time-consuming.~To solve this issue, researchers have began to develop unsupervised MFIF algorithms.~For example, Yan et al.~\cite{yan2018unsupervised} proposed the first unsupervised MFIF algorithm based on CNN, namely MFNet.~The key to achieve unsupervised training in that work was the usage of a loss function based on SSIM, which is a widely used image fusion evaluation metric that measures the structural similarity between source images and the fused image.~Ma et al.~\cite{ma2019sesf} proposed an unsupervised MFIF algorithm based on an encoder-decoder network (SESF), which also utilized SSIM as a part of the loss function.~Other unsupervised methods include DIF-Net~\cite{jung2020unsupervised} and FusionDN~\cite{xu2020fusiondn}.

Apart from CNN, some other deep learning models have also been utilized to perform MFIF.~For example, Guo et al.~\cite{guo2019fusegan} presents the first GAN-based MFIF algorithm (FuseGAN).~Deshmukh et al.~\cite{deshmukh2017multi} proposed to use deep belief network (DBN) to calculate weights indicating the sharp regions of input images.~Unlike above-mentioned methods which only use one model in their methods, Naji et al.~\cite{amin-naji2019ensemble} proposed a MFIF algorithm based on the ensemble of three CNNs.

\section{Multi-focus image fusion benchmark}
\label{sec:benchmark}
As presented previously, in most MFIF works, the algorithm were tested on a small number of images and compared with a very limited number of algorithms using just several evaluation metrics.~This makes it difficult to comprehensively evaluate the real performance of these algorithms.~This section presents a multi-focus image fusion benchmark (MFIFB), including the dataset, baseline algorithms, and evaluation metrics.

\subsection{Dataset}
The dataset in MFIFB is a test set including 105 pairs of multi-focus images.~Each pair consists of two images with different focus areas.~Because most researches in MFIF are about fusing two images, therefore at the moment only image pairs consisting of two images are collected in MFIFB.~Since this paper aims to create a benchmark in the field of MFIF, thus to maximize its value, this test set consists of existing datasets which do not have code library and results.~Specifically, the test set is collected from Lytro \cite{nejati2015multi}, MFFW \cite{xu2020mffw}, the dataset of Savic et al.~\cite{savic2012multifocus}, Aymaz et al.~\cite{aymaz2020multi}, and Tsai et al.~\footnote{https://www.mathworks.com/matlabcentral/fileexchange/45992-standard-images-for-multifocus-image-fusion}.~By doing this, we not only provide benchmark results on the whole dataset, but also give benchmark results for these existing datasets, which will make it more convenient for researchers who are familiar with these datasets to compare results.

The images included in MFIFB are captured with various cameras at various places, and they cover a wide range of environments and working conditions.~The resolutions of images vary from 178$\times$134 to 1024$\times$768.~Therefore, these images can be used to test the performance of MFIF algorithms comprehensively.~Table \ref{table:dataset} lists more details about different kinds of images included in MFIFB.

\begin{table}[H]
	\begin{center}
		\caption{The number of different kinds of images in MFIFB.}
		\label{table:dataset}
		\scriptsize
		\renewcommand\arraystretch{1.5}
		\begin{tabular}{|c|c|c|c|}	
			\hline
			Category  &   Color/gray  & Real/simulated   & Registered/not well registered  \\  \hline
			Number    &   71/34      &  64/41            &  98/7                          \\	    
			\hline
		\end{tabular}
	\end{center}
\end{table}

\subsection{Baseline algorithms}
MFIFB currently contains 30 recently published multi-focus image fusion algorithms including 
ASR \cite{liu2014simultaneous},
BFMF \cite{zhang2017boundary}, BGSC \cite{tian2011multi}, CBF \cite{kumar2015image},  CNN \cite{liu2017multi}, CSR \cite{liu2016image}, DCT\textunderscore Corr \cite{amin2018multi},   
DCT\textunderscore EOL \cite{amin2018multi}, 
DRPL \cite{li2020drpl}, DSIFT \cite{liu2015multi}, DWTDE \cite{liu2013multi},  ECNN \cite{amin-naji2019ensemble}, 
GD \cite{paul2016multi}, GFDF \cite{qiu2019guided},  GFF \cite{li2013image}, IFCNN \cite{zhang2020ifcnn}, 
IFM \cite{li2013imageb}, 
MFM \cite{ma2017multi}, MGFF \cite{bavirisetti2019multi}, 
MST\textunderscore SR \cite{liu2015general}, MSVD \cite{naidu2011image}, MWGF \cite{zhou2014multi}, 
NSCT\textunderscore SR \cite{liu2015general}, 
PCANet \cite{song2018multi}, 
QB \cite{bai2015quadtree},	 	
RP\textunderscore SR \cite{liu2015general}, SESF \cite{ma2019sesf}, SFMD \cite{li2015multi}, SVDDCT \cite{amin2017multi}, TF \cite{ma2019multi}.~In these algorithms, some were specifically designed for multi-focus image fusion, such as ASR and BADNN, while some were designed for general image fusion including multi-focus image fusion, such as CBF and GFF.~It should be noted that some algorithms were originally developed for fusing grayscale images, e.g.~BFMF and CBF.~These algorithms were converted to fuse color images in this study by fusing R, G and B channels, respectively.~More details about the category of the integrated algorithms can be found in Table \ref{table:integrated}.

\begin{table*}
	\begin{center}
		\caption{Multi-focus image fusion algorithms that have been integrated in MFIFB.}
		\label{table:integrated}
		\scriptsize
		\renewcommand\arraystretch{1.5}
		\begin{tabular}{l|p{3cm}|p{10cm}}	
			\hline
			\multicolumn{2}{l|}{Category}         & Method  \\  \hline
			\multicolumn{2}{l|}{Spatial domain-based}      & BFMF \cite{zhang2017boundary}, BGSC \cite{tian2011multi},  DSIFT \cite{liu2015multi},   IFM \cite{li2013imageb}, QB \cite{bai2015quadtree}, TF \cite{ma2019multi}   \\ \hline 
			\multirow{2}{*}{Transform domain-based} & SR-based & ASR \cite{liu2014simultaneous}, CSR \cite{liu2016image} \\  \cline{2-3} 
			                                        & Multi-scale-based & CBF \cite{kumar2015image}, DCT\textunderscore Corr \cite{amin2018multi}, 
			                                        DCT\textunderscore EOL \cite{amin2018multi},  
			                                        DWTDE \cite{liu2013multi}, GD \cite{paul2016multi}, MSVD \cite{naidu2011image}, MWGF \cite{zhou2014multi},  SVDDCT \cite{amin2017multi}     \\ \cline{2-3}
			                                        & edge-preserving-based& GFDF \cite{qiu2019guided}, GFF \cite{li2013image}, MFM \cite{ma2017multi}, MGFF \cite{bavirisetti2019multi}   \\ \cline{2-3}                          			                                         
													&subspace-based      &   SFMD \cite{li2015multi}    \\  \hline 
			\multicolumn{2}{l|}{Hybrid}      & MST\textunderscore SR \cite{liu2015general}, NSCT\textunderscore SR \cite{liu2015general}, RP\textunderscore SR \cite{liu2015general}\\	     \hline
			\multirow{2}{*}{Deep learning-based}    & Supervised      &CNN \cite{liu2017multi}, DRPL \cite{li2020drpl}, ECNN \cite{amin-naji2019ensemble}, IFCNN \cite{zhang2020ifcnn}, PCANet \cite{song2018multi}  \\	 \cline{2-3} 
			                                        & Unsupervised  &SESF \cite{ma2019sesf} \\  
			\hline
		\end{tabular}
	\end{center}
\end{table*}

\begin{figure*}
	\centering
	\includegraphics[width=\textwidth]{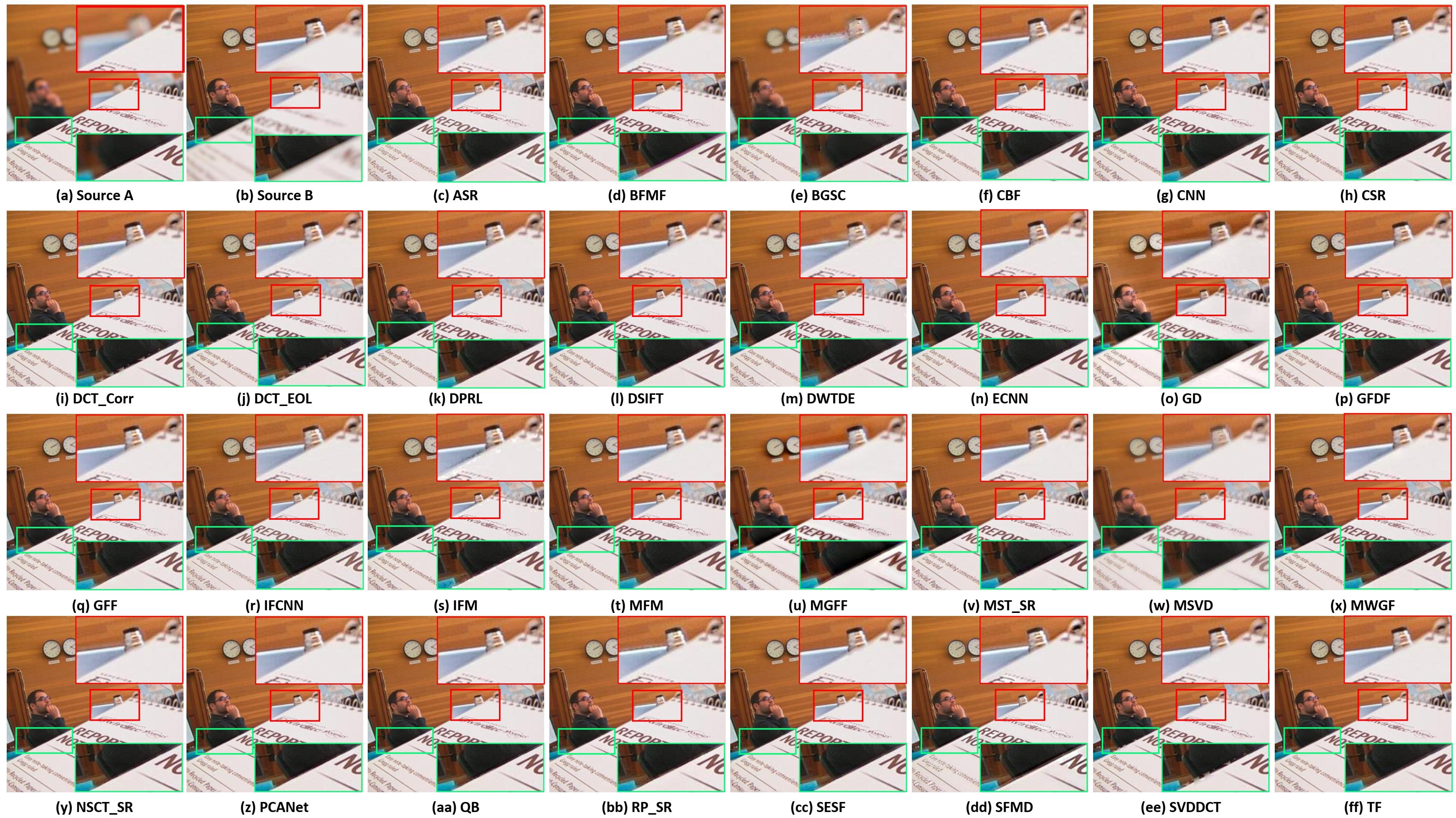}
	\caption{The source images and fused images of \textit{Lytro19} image pair.~(a) and (b) are the source images.~From (c) to (ff) are the fused images produced by 30 integrated MFIF algorithms in MFIFB.~The magnified plot of area within red box near the focused/defocused boundary are given at the top right corner of each fused image.~The magnified plot of area within green box near the focused/defocused boundary are given at the bottom right corner of each fused image.}
	\label{fig:qualitative-lytro19}
\end{figure*}

\begin{figure*}
	\centering
	\includegraphics[width=\textwidth]{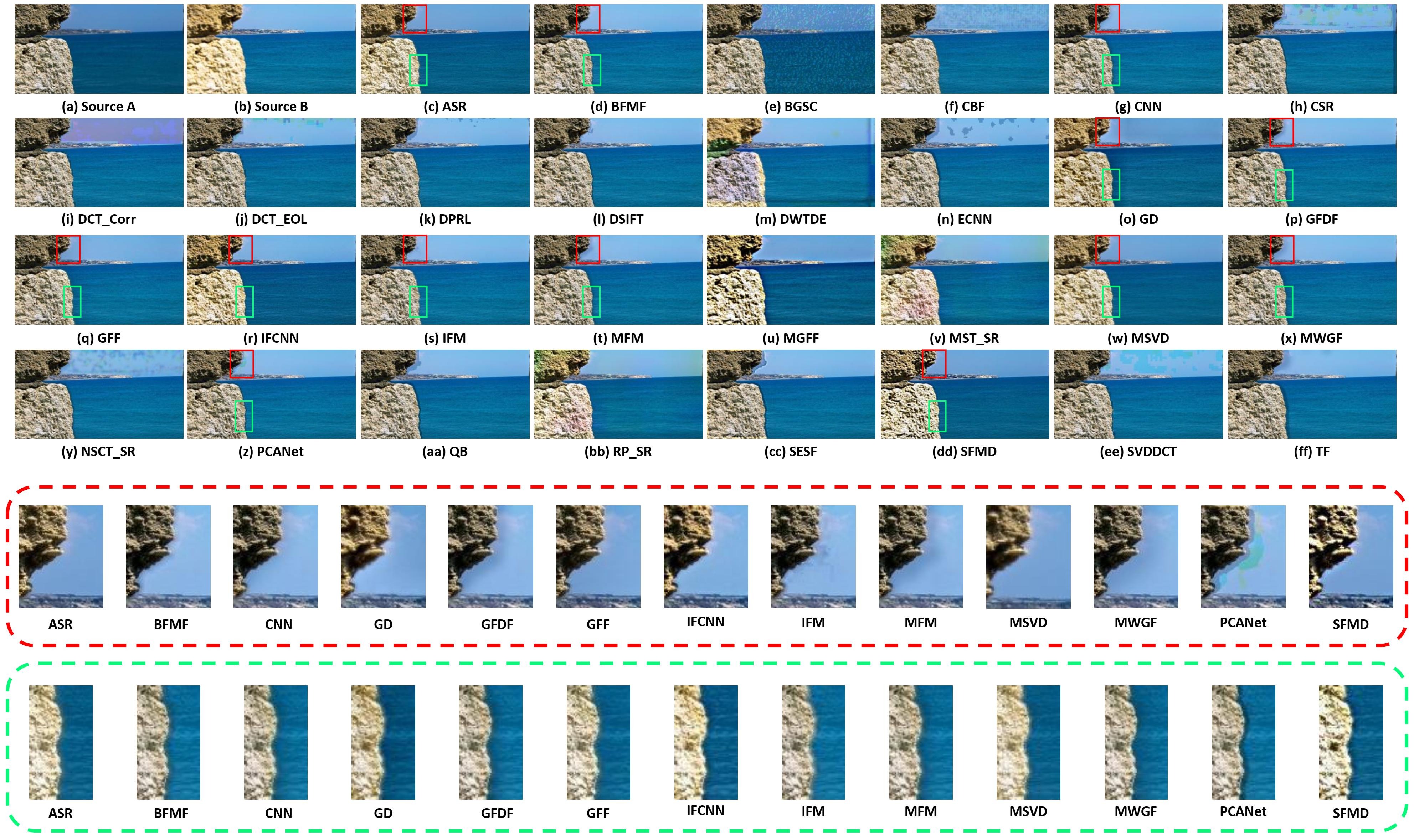}
	\caption{The source images and fused images of \textit{MMFW12} image pair.~(a) and (b) are the source images.~From (c) to (ff) are the fused images produced by 30 integrated MFIF algorithms in MFIFB.~The magnified plot of area within red box near the focused/defocused boundary are given in the red dash box.~The magnified plot of area within green box near the focused/defocused boundary are given in the green dash box.}
	\label{fig:qualitative-mmfw12}
\end{figure*}

The algorithms in MFIFB cover almost every kind of MFIF algorithms,  
thus can represent the development of the field to some extent.~However, it should be noted that only a part of published MFIF methods provide the source code, thus in MFIFB we cannot cover all published MFIF algorithms.

The source codes of various methods have different input and output interfaces, and they may require different running environment.~These factors hinder the usage of these codes to produce results and compare performances.~To integrate algorithms into MFIFB and for the convenience of users,  an interface was designed to integrate more algorithms into MFIFB.~Besides, for researchers who do not want to make their codes publicly available, they can simply put their fused images into MFIFB and then their algorithms can be compared with those integrated in MFIFB easily.

\subsection{Evaluation metrics}
The assessment of MFIF algorithms is not a trivial task since the ground-truth images are normally not available.~Generally there are two ways to evaluate MFIF algorithms, namely subjective or qualitative method and objective or quantitative method.

Subjective evaluation means that the fusion performance is evaluated by human observers.~This is very useful in MFIF research since a good fused image should be friendly to human visual system.~However, it is time-consuming and labor-intensive to observe each fused image in practice.~Besides, because each observer has different standard when observing fused images, thus biased evaluation may be easily produced.~Therefore, qualitative evaluation alone is not enough for the fusion performance evaluation.~Therefore, objective evaluation metrics are needed for quantitative comparison.

As introduced in \cite{liu2012objective}, image fusion evaluation metrics can be classified into four types as
\begin{itemize}
	\item Information theory-based
	
	\item Image feature-based
	
	\item Image structural similarity-based
	
	\item Human perception inspired
\end{itemize}

Numerous evaluation metrics for image fusion have been proposed.~However, none of them is better than all other metrics.~To have comprehensive and objective performance comparison, 20 evaluation metrics were implemented in MFIFB \footnote{The implementation of some metrics are kindly provided by Zheng Liu at https://github.com/zhengliu6699/imageFusionMetrics}.~The evaluation metrics integrated in MFIFB cover all four categories of metrics, thus are capable of quantitatively showing the quality of a fused image.~Specifically, the implemented information theory-based metrics include 
cross entropy (CE) \cite{bulanon2009image}, entropy (EN) \cite{aardt2008assessment}, feature mutual information (FMI), normalized mutual information (NMI) \cite{hossny2008comments},  peak signal-to-noise ratio (PSNR) \cite{jagalingam2015review}, nonlinear correlation information entropy (Q$_{NCIE}$)  \cite{wang2005nonlinear, wang2008performance}, and tsallis entropy (TE) \cite{cvejic2006image}.~The implemented image feature-based metrics include average gradient (AG) \cite{cui2015detail}, edge intensity (EI) \cite{rajalingam2018hybrid}, gradient-based similarity measurement ($Q^{AB/F}$) \cite{xydeas2000objective}, phase congruency ($Q_P$) \cite{zhao2007performance}, standard division (SD)  \cite{rao1997fibre} and spatial frequency (SF) \cite{eskicioglu1995image}.~The implemented image structural similarity-based metrics include Cvejie's metric $Q_C$ \cite{cvejic2005similarity}, Peilla's metric ($Q_W$) \cite{piella2003new},  Yang's metric $(Q_Y)$ \cite{yang2008novel}, and structural similarity index measure (SSIM) \cite{wang2004image}.~The implemented human perception inspired fusion metrics are human visual perception ($Q_{CB}$) \cite{chen2009new}, $Q_{CV}$ \cite{chen2007human} and VIF \cite{han2013new}.

Due to the page limitation, the mathematical expression of these metrics are not given here.~For all metrics except CE and  $Q_{CV}$, a larger value indicates a better fusion performance.~In MFIFB, it is convenient to compute all these metrics for each method, making it easy to compare performances.~Note that many metrics are designed for gray images.~In this work, each metric was computed for every channel of RGB images and then the average value was computed.~More information about evaluation metrics can be founded in \cite{liu2012objective, jagalingam2015review, ma2019infrared}.

\begin{sidewaystable}	
	\begin{center}
		\caption{Average evaluation metric values of all methods on the Lytro dataset (20 image pairs).~The best three values in each metric are denoted in \xz{red}, \xzg{green} and \xzb{blue}, respectively.~The three numbers after the name of each method denote the number of best value, second best value and third best value, respectively.~Best viewed in color.} 
		\label{table:metrics_average-lytro}
		\tiny
		\begin{tabular*}{\textwidth}{@{}@{\extracolsep{\fill}}llllllll|llllll|llll|lll@{}}	
			\hline
			Method      & CE  & EN  & FMI  & NMI  & PSNR  &  $Q_{NCIE}$ & TE  & AG  & EI & $Q^{AB/F}$ & $Q_P$  & SD  & SF &$Q_C$ & $Q_W$& $Q_Y$ & SSIM & $Q_{CB}$ & $Q_{CV}$ & VIF    \\ \hline
			BFMF (0,1,0) & 0.019638 & 7.524990 & 0.899890 & 1.088442 & 64.613500 & \xzg{0.839420} & 77.423500 & 6.843840 & 71.110150 & 0.739521 & 0.813292 & 61.329300 & 19.183530 & 0.809169 & 0.928667 & 0.972186 & 1.669885 & 0.791523 & 22.259420 & 0.924554 \\
			
			BGSC (1,0,0) & \xz{0.015643} & 7.503900 & 0.882147 & 0.924923 & 64.813000 & 0.831577 & 46.645900 & 5.407235 & 55.588700 & 0.517730 & 0.463801 & 59.794650 & 15.942770 & 0.754699 & 0.589341 & 0.847103 & 1.643050 & 0.655019 & 131.57760 & 0.757423 \\
			
			DSIFT (0,2,2) & 0.020353 & 7.526410 & 0.899970 & \xzg{1.097098} & 64.542300 & \xzb{0.839410} & 78.613300 & \xzb{6.937035} & \xzg{72.050850} & 0.744813 & 0.818027 & 61.583800 & 19.486470 & 0.811880 & 0.946815 & 0.974128 & 1.667965 & \xzb{0.796375} & 16.751350 & 0.937378 \\
			
			IFM (0,0,0)& 0.020409 & 7.528530 & 0.897394 & 1.068791 & 64.497000 & 0.837834 & 73.180550 & 6.931470 & 71.971200 & 0.736289 & 0.793869 & 61.512050 & 19.455460 & 0.805784 & 0.937143 & 0.966946 & 1.662950 & 0.785079 & 21.312070 & 0.930432 \\
			
			QB (1,0,3) & 0.020324 & 7.526225 & 0.899898 & \xzb{1.096909} & 64.542850 & 0.839371 & \xzb{78.664450} & 6.929265 & 71.970600 & 0.744117 & 0.815647 & 61.585100 & 19.482670 & 0.811833 & 0.945604 & \xzb{0.974253} & 1.668035 & \xz{0.796537} & 17.094340 & 0.936397 \\
			
			TF (2,3,0) & 0.020177 & 7.526195 & 0.899791 & 1.090800 & 64.572850 & 0.838992 & 77.612350 & 6.908155 & 71.751500 & \xz{0.745418} & \xzg{0.819273} & 61.553250 & 19.409010 & \xz{0.812919} & \xzg{0.947745} & \xzg{0.974405} & 1.670550 & 0.795914 & 16.820720 & 0.937493 \\ \hline 
			
			ASR (0,0,2) & \xzb{0.019154} & 7.527545 & 0.898110 & 0.906832 & \xzb{64.829750} & 0.828414 & 37.959950 & 6.778255 & 70.215350 & 0.724654 & 0.785958 & 60.924050 & 19.067270 & 0.801011 & 0.943607 & 0.949396 & 1.671965 & 0.710252 & 27.033780 & 0.897371 \\
			
			CSR (0,1,0) & 0.020296 & 7.527975 & 0.899603 & 1.030608 & 64.578600 & 0.835017 & 62.038200 & 6.839740 & 71.083900 & 0.733271 & 0.806346 & 61.490100 & 19.321700 & 0.794129 & 0.946077 & 0.944537 & 1.670385 & 0.768573 & \xzg{16.548670} & 0.932877 \\ \hline 
			
			CBF (0,2,0) & 0.019302 & 7.530505 & 0.897206 & 0.976413 & \xzg{64.853300} & 0.832134 & 48.834550 & 6.726525 & 69.760700 & 0.732017 & 0.793537 & 60.907850 & 18.676130 & 0.808502 & 0.933265 & 0.953732 & \xzg{1.679230} & 0.749864 & 26.241540 & 0.914540 \\
			
			DCT\_Corr (0,0,0) & 0.020061 & 7.527230 & 0.899367 & 1.089727 & 64.547850 & 0.839064 & 76.490450 & 6.922705 & 71.901750& 0.740326 & 0.805902 & 61.501650 & 19.426950 & 0.809230 & 0.934770 & 0.968633 & 1.667025 & 0.787151 & 21.457040 & 0.930913 \\
			
			DCT\_EOL (0,0,0) & 0.020249 & 7.527355 & 0.899547 & 1.094470 & 64.541650 & 0.839282 & 77.509100 & 6.932225 & 72.000900 & 0.742283 & 0.810574 & 61.552050 & 19.460770 & 0.810006 & 0.938988 & 0.969555 & 1.667215 & 0.789194 & 20.117330 & 0.933844 \\
			
			DWTDE (0,1,0) & \xzg{0.018957} & 7.527975 & 0.898809 & 1.006858 & 64.606700 & 0.833793 & 61.730100 & 6.752600 & 70.104950 & 0.716354 & 0.784280 & 61.254400 & 18.920690 & 0.802234 & 0.932183 & 0.952613 & 1.667810 & 0.767808 & 29.083500 & 0.918895 \\
			
			\xzb{GD (3,0,0)} & 0.396174 & \xz{7.607335} & 0.887266 & 0.511936 & 61.272450 & 0.813712 & \xz{275.11160} & 6.890210 & 71.746800 & 0.677429 & 0.693197 & 57.306550 & 18.024320 & 0.732460 & 0.869108 & 0.838173 & 1.557465 & 0.588832 & 127.11860 & \xz{1.000538} \\
			
			MSVD (2,0,0) & 0.022366 & 7.497410 & 0.879994 & 0.789597 & \xz{65.247500} & 0.822698 & 44.968990 & 4.803970 & 49.542950 & 0.500233 & 0.518228 & 58.954500 & 14.433330 & 0.724487 & 0.704862 & 0.790878 & \xz{1.692465} & 0.604281 & 86.149790 & 0.748725 \\
			
			MWGF (1,0,0) & 0.020377 & 7.527275 & \xz{0.900344} & 1.065568 & 64.540150 & 0.837937 & 71.925800 & 6.815705 & 70.876900 & 0.732472 & 0.809721 & 61.469200 & 19.257570 & 0.809685 & 0.929381 & 0.973900 & 1.668205 & 0.785379 & 19.699000 & 0.928065 \\
			
			SVDDCT (0,0,0) & 0.020117 & 7.526470 & 0.899656 & 1.094793 & 64.545150 & 0.839381 & 78.376900 & 6.927960 & 71.954200 & 0.742519 & 0.811085 & 61.555000 & 19.465680 & 0.806462 & 0.941773 & 0.969724 & 1.667475 & 0.788470 & 18.025000 & 0.934466 \\ \hline 
			
			GFDF (2,2,3) & 0.020140 & 7.526160 & 0.899992 & 1.090750 & 64.573150 & 0.838961 & 77.344550 & 6.902755 & 71.700900 & \xzb{0.745061} & \xz{0.819468} & 61.549350 & 19.397050 & \xzg{0.812899} & \xzb{0.947740} & \xz{0.974458} & 1.670630 & \xzg{0.796394} & \xzb{16.640480} & 0.937175 \\
			
			GFF (0,0,0) & 0.020287 & 7.529880 & 0.899365 & 1.040993 & 64.612600 & 0.835995 & 64.040550 & 6.891830 & 71.552850 & 0.742155 & 0.815251 & 61.463850 & 19.347270 & 0.810463 & 0.946958 & 0.969002 & 1.670730 & 0.782379 & 17.139130 & 0.934393 \\
			
			MFM (0,1,0) & 0.020050 & 7.526425 & 0.899556 & 1.088707 & 64.584700 & 0.838992 & 76.910750 & 6.898890 & 71.653400& \xzg{0.745218} & 0.817426 & 61.517200 & 19.368160 & 0.812455 & 0.945857 & 0.974237 & 1.670915 & 0.794728 & 17.405300& 0.935602 \\
			
			MGFF (1,1,1) & 0.046011 & \xzb{7.534860} & 0.887996 & 0.766430 & 64.010550 & 0.821890 & 28.079500 & 6.187375 & 64.494550 & 0.644779 & 0.677019 & \xz{63.218000} & 17.443950 & 0.773243 & 0.887198 & 0.872198 & 1.671740 & 0.653263 & 381.75990 & \xzg{0.975855} \\ \hline 
			
			\xz{SFMD (3,2,1)} & 0.031383 & \xzg{7.535560} & 0.886008 & 0.759986 & 63.594700 & 0.821765 & 24.306030 & \xz{8.153865} & \xz{84.403400} & 0.640447 & 0.680628 & \xzg{62.995500} & \xz{23.409000} & 0.772858 & 0.889009 & 0.896568 & 1.624320 & 0.636672 & 77.419900 & \xzb{0.949093} \\ \hline 
			
			MST\_SR (1,0,1) & 0.022262 & 7.528345 & 0.899067 & 0.952416 & 64.601550 & 0.830625 & 43.776450 & 6.932290 & 71.998900 & 0.735014 & 0.806941 & \xzb{61.765850} & 19.445200 & 0.805785 & \xz{0.947812} & 0.956312 & 1.669545 & 0.756733 & 19.617030 & 0.946196 \\
			
			NSCT\_SR (0,0,0) & 0.020021 & 7.528515 & 0.899531 & 1.059497 & 64.612850 & 0.837055 & 69.128800 & 6.908970 & 71.736300 & 0.741461 & 0.814431 & 61.438350 & 19.399950 & 0.810613 & 0.947641 & 0.966277 & 1.671110 & 0.781873 & 16.895400 & 0.934844 \\
			
			RP\_SR (0,1,0) & 0.022805 & 7.529890 & 0.895937 & 0.925847 & 64.607100 & 0.829275 & 38.968250 & 6.928545 & 71.892200 & 0.720292 & 0.776160 & 61.649750 & \xzg{19.586560} & 0.803218 & 0.940219 & 0.947480 & 1.671150 & 0.738752 & 20.643050 & 0.939251 \\ \hline 
			
			CNN (0,0,3) & 0.019920 & 7.526245 & 0.899765 & 1.082380 & 64.602850 & 0.838458 & 75.472550 & 6.875200 & 71.408200 & 0.744251 & \xzb{0.818436} & 61.495350 & 19.298950 & \xzb{0.812681} & 0.946088 & 0.973324 & \xzb{1.672075} & 0.794012 & 17.108970 & 0.935537 \\
			
			DPRL (0,1,2) & 0.020867 & 7.528040 & 0.899082 & 1.017829 & 64.549050 & 0.834265 & 56.287000 & \xzg{6.939190} & \xzb{72.024350} & 0.737734 & 0.817047 & 61.591950 & \xzb{19.519400} & 0.810174 & 0.946562 & 0.971286 & 1.668705 & 0.773679 & 17.734790 & 0.937124 \\
			
			\xzg{ECNN (3,2,0)} & 0.020178 & 7.525930 & \xzg{0.900061} & \xz{1.101154} & 64.536700 & \xz{0.839672} & \xzg{80.749550} & 6.911405 & 71.755850 & 0.741015 & 0.805103 & 61.580400 & 19.476880 & 0.810527 & 0.944447 & 0.970987 & 1.667370 & 0.792449 & \xz{16.262140} & 0.934347 \\
			
			IFCNN (0,0,0) & 0.037096 & 7.530950 & 0.896192 & 0.882476 & 64.645100 & 0.826998 & 37.368850 & 6.933880 & 71.989350 & 0.712780 & 0.771835 & 61.444050 & 19.442410 & 0.803788 & 0.942935 & 0.938947 & 1.670895 & 0.705921 & 19.832180 & 0.933636 \\
			
			PCANet (0,0,1) & 0.020317 & 7.524715 & \xzb{0.900045} & 1.087111 & 64.531900 & 0.838525 & 76.624150 & 6.880480 & 71.471500 & 0.738438 & 0.814058 & 61.550950 & 19.395400 & 0.810176 & 0.939627 & 0.973351 & 1.667850 & 0.792703 & 16.999560 & 0.934086 \\
			
			SESF (0,0,0) & 0.020269 & 7.526615 & 0.899455 & 1.081988 & 64.541250 & 0.838440 & 77.493900 & 6.909530 & 71.788650 & 0.739873 & 0.811905 & 61.588100 & 19.456790 & 0.809254 & 0.946288 & 0.971380 & 1.668120 & 0.793744 & 18.023750 & 0.938644 \\
			\hline
		\end{tabular*}
	\end{center}
\end{sidewaystable}	

\begin{sidewaystable}	
	\begin{center}
		\caption{Average evaluation metric values of all methods on the whole MFIFB dataset (105 image pairs).~The best three values in each metric are denoted in \xz{red}, \xzg{green} and \xzb{blue}, respectively.~The three numbers after the name of each method denote the number of best value, second best value and third best value, respectively.~Best viewed in color.} 
		\label{table:metrics_average}
		\tiny
		\begin{tabular*}{\textwidth}{@{}@{\extracolsep{\fill}}llllllll|llllll|llll|lll@{}}	
			\hline
			Method      & CE  & EN  & FMI  & NMI  & PSNR  &  $Q_{NCIE}$ & TE  & AG  & EI & $Q^{AB/F}$ & $Q_P$  & SD  & SF &$Q_C$ & $Q_W$& $Q_Y$ & SSIM & $Q_{CB}$ & $Q_{CV}$ & VIF    \\ \hline
			BFMF (0,1,2)  & 0.047329 & 7.180948 & 0.897034 & \xzb{1.111096} & 63.332440 & \xzb{0.843143} & 317.580100 & 7.609427 & 74.920800 & 0.743772 & 0.788129 & 56.235860 & 23.330390 & 0.816095 & 0.906927 & \xzg{0.949828} & 1.663249 & 0.787603 & 118.57960 & 0.867944 \\
			
			BGSC (1,0,1) & \xz{0.037523} & 7.173506 & 0.884569 & 1.008714 & \xzb{63.505650} & 0.841565 & 220.513200 & 6.658554 & 64.939570 & 0.589086 & 0.508681 & 54.880960 & 21.036710 & 0.773571 & 0.646012 & 0.835154 & 1.622742 & 0.684642 & 214.43930 & 0.746569 \\
			
			DSIFT (0,0,0) & 0.042452 & 7.182328 & 0.896655 & 1.109905 & 63.293060 & 0.842911 & 207.333100 & 7.710156 & 75.839980 & 0.747565 & 0.784749 & 56.323670 & 23.644280 & 0.814340 & 0.917656 & 0.943182 & 1.658877 & 0.785830 & 78.216470 & 0.872679 \\
			
			IFM (1,0,0) & 0.053092 & 7.185199 & 0.895362 & 1.084493 & 63.276100 & 0.841046 & \xz{8824.74500} & 7.673120 & 75.426660 & 0.741902 & 0.778086 & 56.299770 & 23.584020 & 0.814678 & 0.912465 & 0.942871 & 1.661107 & 0.770410 & 81.069980 & 0.870643 \\
			
			\xz{QB (4,1,1)} & 0.045462 & 7.179638 & 0.896987 & \xz{1.117892} & 63.284990 & \xz{0.843277} & 286.881300 & 7.712189 & 75.787760 & \xzg{0.748466} & 0.788975 & 56.390070 & 23.677010 & \xzb{0.817610} & 0.918305 & \xz{0.950054} & 1.660263 & \xz{0.789534} & 113.74240 & 0.872385 \\
			
			TF (0,0,1) & 0.044351 & 7.180714 & 0.897006 & 1.100021 & 63.319040 & 0.842044 & 238.296300 & 7.656085 & 75.287980 & 0.747829 & \xzb{0.791328} & 56.308030 & 23.537160 & 0.816977 & 0.921313 & 0.944670 & 1.665278 & 0.786408 & 71.291770 & 0.875204 \\ \hline 
			
			ASR (1,1,0) & 0.051878 & 7.195639 & 0.896024 & 0.941034 & \xz{63.647060} & 0.832749 & 825.225200 & 7.495725 & 73.495640& 0.727783 & 0.761004 & 55.717590 & 23.147130 & 0.815282 & 0.917409 & 0.921470 & \xzg{1.673467} & 0.724397 & 74.219540 & 0.848602 \\
			
			CSR (0,0,0) & 0.063374 & 7.212798 & 0.895318 & 0.926963 & 63.344380 & 0.831827 & 324.279900 & 7.431270 & 73.575560 & 0.719437 & 0.763687 & 56.139080 & 23.000090 & 0.782021 & 0.918944 & 0.893531 & 1.661639 & 0.741709 & 71.735620 & 0.863496 \\ \hline 
			
			CBF (1,1,1) & 0.046291 & 7.190175 & 0.894211 & 0.992856 & \xzg{63.606440} & 0.836229 & 135.397800 & 7.373120 & 72.618800 & 0.730552 & 0.749472 & 55.492600 & 22.449820 & 0.812668 & 0.910816 & 0.910066 & \xz{1.679255} & 0.745514 & \xzb{70.333860} & 0.864294 \\
			
			DCT\_Corr (0,0,0) & 0.054290 & 7.183771 & 0.894585 & 1.108963 & 63.114500 & 0.842871 & 322.045100 & 7.708113 & 75.921080 & 0.739957 & 0.760543 & 56.216510 & 23.710990 & 0.812211 & 0.907921 & 0.938217 & 1.655281 & 0.772510 & 116.74420 & 0.867278 \\
			
			DCT\_EOL (0,1,0) & 0.047559 & 7.183358 & 0.894729 & \xzg{1.111800} & 63.113930 & 0.842919 & 222.824100 & 7.721065 & 76.048910 & 0.741991 & 0.764724 & 56.321780 & 23.743720 & 0.812070 & 0.912016 & 0.938446 & 1.655007 & 0.773732 & 114.02590 & 0.869921 \\
			
			DWTDE (0,0,0) & 0.044099 & 7.191523 & 0.894229 & 1.002405 & 63.407360 & 0.837050 & 1610.87700 & 7.257119 & 72.359080 & 0.709249 & 0.739962 & 55.855390 & 21.644490 & 0.803161 & 0.895561 & 0.909535 & 1.662259 & 0.757619 & 106.96460 & 0.861150 \\
			
			GD (2,2,0) & 0.836936 & \xz{7.478091} & 0.882739 & 0.518788 & 59.720350 & 0.816733 & 443.025500 & \xzg{8.283411} & \xzg{81.765120} & 0.674440 & 0.679038 & 56.279950 & 23.674930 & 0.726548 & 0.839127 & 0.798234 & 1.527683 & 0.601711 & 159.64610 & \xz{1.108664} \\
			
			MSVD (0,0,0) & 0.185617 & 7.185232 & 0.880630 & 0.821143 & 63.463020 & 0.826806 & 3572.78600& 5.847351 & 57.158190 & 0.562932 & 0.562638 & 53.741960 & 18.796280 & 0.740471 & 0.727040 & 0.784893 & 1.662920 & 0.626127 & 128.50170 & 0.751377 \\
			
			MWGF (0,1,0) & 0.052617 & 7.193130 & \xzb{0.897046} & 1.056287 & 63.253240 & 0.839154 & 1565.94100 & 7.604590 & 74.893540 & 0.737120 & 0.786354 & 56.332190 & 23.434040 & 0.815298 & 0.914490 & 0.946989 & 1.661468 & 0.777470 & 116.48420 & 0.875608 \\
			
			SVDDCT (0,0,0) & 0.049238 & 7.183745 & 0.894715 & 1.110505 & 63.116950 & 0.842871 & 318.700000 & 7.709491 & 75.936780 & 0.740761 & 0.763561 & 56.328190 & 23.710590 & 0.809084 & 0.910232 & 0.938168 & 1.655304 & 0.773117 & 119.41740 & 0.869850 \\ \hline 
			
			\xzb{GFDF (3,2,1)} & 0.044624 & 7.181165 & \xz{0.897068} & 1.104549 & 63.321870 & 0.842246 & 199.680400 & 7.656499 & 75.316880 & \xzb{0.748381} & \xz{0.792594} & 56.291310 & 23.511940 & \xzg{0.818467} & \xz{0.921729} & 0.947203 & 1.665535 &\xzg{0.788212} & 71.242840 & 0.876088 \\
			
			GFF (0,0,0) & 0.083664 & 7.237063 & 0.889518 & 0.826873 & 63.227980 & 0.827092 & 233.429800 & 7.638879 & 75.163040 & 0.678299 & 0.692559 & 56.175990 & 23.345510 & 0.749023 & 0.910094 & 0.856117 & 1.634412 & 0.724829 & 70.534500 & 0.860479 \\
			
			MFM (0,0,0) & 0.050854 & 7.181246 & 0.896919 & 1.094574 & 63.328970 & 0.841762 & 264.700300 & 7.642581 & 75.150600 & 0.746994 & 0.790849 & 56.283410 & 23.493350 & 0.817410 & 0.920185 & 0.943908 & 1.666019 & 0.784931 & 71.491590 & 0.874145 \\
			
			MGFF (0,2,1) & 0.331531 & \xzb{7.284876} & 0.887729 & 0.845459 & 61.976560 & 0.828354 & 639.601900 & 5.843658 & 58.833030 & 0.666799 & 0.684678 & \xzg{60.150050} & 22.440790 & 0.591809 & 0.859168 & 0.835316 & 1.638539 & 0.678664 & 333.73940 & \xzg{1.042627} \\  \hline 
			
			\xzg{SFMD (4,1,1)} & 0.142828 & \xzg{7.285385} & 0.872729 & 0.673841 & 61.686120 & 0.821230& 292.684100 & \xz{10.78023} & \xz{103.94860} & 0.558764 & 0.597324 & \xz{60.499420} & \xz{33.832340} & 0.729447 & 0.803434 & 0.809958 & 1.561684 & 0.602256 & 180.40710 & \xzb{0.943558} \\ \hline 
			
			MST\_SR (0,1,2) & 0.056093 & 7.195381 & 0.895229 & 0.979681 & 63.369550 & 0.835584 & 1749.29400 & 7.748169 & 76.077150 & 0.732042 & 0.762195 & \xzb{56.701530} & 23.699760 & 0.813895 & \xzb{0.921444} & 0.920897 & 1.667245 & 0.753582 & \xzg{70.312950} & 0.898205 \\
			
			NSCT\_SR (1,1,1) & 0.048941 & 7.194330 & 0.896178 & 1.044676 & 63.379310 & 0.839234 & \xzb{3926.16700} & 7.678417 & 75.363690 & 0.739404 & 0.774890 & 56.130330 & 23.577820 & 0.813787 & \xzg{0.921544} & 0.925396 & 1.667406 & 0.765249 & \xz{69.466900} & 0.875830 \\
			
			RP\_SR (0,1,1) & 0.058435 & 7.194332 & 0.892773 & 0.962425 & 63.365850 & 0.834731 & 1111.90900 & \xzb{7.778523} & 76.013000 & 0.718907 & 0.734587 & 56.444050 & \xzg{24.112460} & 0.810809 & 0.911139 & 0.911492 & 1.666139 & 0.738713 & 83.209500 & 0.883536 \\ \hline 
			
			CNN (0,3,0) & \xzg{0.042012} & 7.178280 & \xzg{0.897048} & 1.097481 & 63.337300 & 0.841981 & 202.594000 & 7.615910 & 74.917820 & 0.745743 & \xzg{0.791459} & 56.403500 & 23.412440 & 0.817460 & 0.919990 & 0.946716 & 1.667055 & 0.785701 & 72.016260 & 0.874574 \\
			
			DPRL (2,0,2) & 0.042576 & 7.184383 & 0.895820 & 1.079219 & 63.299200 & 0.839579 & 232.433700 & 6.129938 & 76.425780 & \xz{0.748762} & 0.782818 & 56.439470 & \xzb{24.073080} & \xz{0.823692} & 0.916191 & 0.947460 & \xzb{1.668405} & 0.776957 & 80.815330& 0.881630\\
			
			ECNN (0,0,1) & \xzb{0.042090} & 7.184773 & 0.893811 & 1.009942 & 63.222810 & 0.837058 & 126.762800 & 7.729687 & 75.898300 & 0.716196 & 0.735345 & 56.371520 & 23.690570 & 0.786739 & 0.912133 & 0.900764 & 1.645200 & 0.759885 & 75.727840 & 0.865427 \\
			
			IFCNN (0,0,1) & 0.115571 & 7.235051 & 0.888923 & 0.826030 & 63.428250 & 0.826721 & 194.994000 & 6.037408 & \xzb{76.718000} & 0.684346 & 0.694605 & 56.378290 & 23.608870 & 0.785120 & 0.904506 & 0.870977 & 1.666035 & 0.697467 & 74.739380 & 0.887171 \\
			
			PCANet (0,1,2) & 0.046250 & 7.180769 & 0.896946 & 1.110546 & 63.336390 & \xzg{0.843242} & 218.566100 & 7.686064 & 75.543250 & 0.743863 & 0.785662 & 56.327740 & 23.574830 & 0.813666 & 0.910947 & \xzb{0.948250} & 1.660594 & \xzb{0.787905} & 118.87970 & 0.870008 \\
			
			SESF (0,1,0) & 0.049232 & 7.185491 & 0.895610 & 1.085557 & 63.275910 & 0.841288 & \xzg{8614.76300} & 7.688576 & 75.702570 & 0.741529 & 0.780424 & 56.385010 & 23.674230 & 0.810245 & 0.918252 & 0.941403 & 1.658370 & 0.779072 & 86.897670 & 0.874885 \\
			\hline 
		\end{tabular*}
	\end{center}
\end{sidewaystable}	

\section{Experiments and analysis}
\label{sec:experiment}

This section presents experimental results within MFIFB.~All experiments were performed using a computer equipped with an NVIDIA RTX2070 GPU and i7-9750H CPU.~Default parameters reported by the corresponding authors of each algorithm were employed.~Note that pre-trained models of each deep learning algorithm were provided by the corresponding authors of each algorithm.~The dataset in MFIFB is only used for performance evaluation  of those algorithms but not for the training.

\subsection{Results on the Lytro dataset}
\label{subsec:lytro}
Many papers utilize Lytro in the experiments, thus the Lytro dataset is collected as a subset in MFIFB and in this Section the results on the Lytro dataset are presented.
\subsubsection{Qualitative performance comparison}
Figure \ref{fig:qualitative-lytro19} illustrates the fused images of all integrated algorithms in MFIFB on the \textit{Lytro19} image pair.~As can be seen, most algorithms can produce a clear image in this case while the BGSC and MSVD give blurring ones.~To further investigate the focused/defocused boundary in the fused images, two magnified plots are given in Fig.~\ref{fig:qualitative-lytro19} for each image.~As can be seen, many algorithms cannot well handle the boundary area contained in the red box, including ASR, BGSC, CBF, DWTDE, GD, GFF, IFCNN, IFM, MFM, MGFF, MSVD, NSCT\_SR, RP\_SR, SFMD.~Besides, some algorithms cannot fuse the boundary area contained in the green box well, including BFMF, BGSC, CBF, DCT\_Corr, DCT\_EOL, DWTDE, GD, IFCNN, IFM, MSVD, SFMD, SVDDCT.~To sum up, the remaining methods, namely CNN, CSR, DPRL, DSIFT, ECNN, GFDF, MSR\_SR, MWGF, PCANet, QB, SESF, and TF, have similar visual performances on the \textit{Lytro19} image pair.~Among these methods, five are deep learning-based methods, QB and TF are spatial domain-based methods while the rest are transform domain-based ones.

\subsubsection{Quantitative performance comparison}
Table \ref{table:metrics_average-lytro} lists the average value of 20 evaluation metrics for all methods on the Lytro dataset.~As can be seen, the top three methods are SFMD, ECNN and GD, respectively.~Specifically, GD and SFMD are transform domain-based methods while ECNN is a deep learning-based approach.~This means that the transform domain-based methods achieve the best results on the Lytro dataset, and deep learning-based methods also obtain competitive results.~Note that although these three methods all have the best value in three evaluation metrics, they show different characteristics.~To be more specific, SFMD only performs well in image feature-based metrics, while ECNN and GD exhibit good performances in both information theory-based and human perception inspired metrics.

Actually, from the table one can find more about the performance of each kind of methods.~First, the spatial domain-based approaches do not show competitive performances except TF.~In transform domain-based methods, SR-based ones perform poorly in most metrics.~Multi-scale-based approaches have better performance in information theory-based methods while edge-preserving-based algorithms generally perform better in image feature-based metrics.~Similar to SR-based ones, the hybrid methods which combines multi-scale and SR approaches do not show good performance in most metrics.~Regarding deep learning-based methods, although ECNN ranks the second among all 30 algorithms, other deep learning-based methods do not perform well.~This is supervising because deep learning can provide good features and can learn fusion rules automatically.~This may because that most deep learning-based algorithms were trained using simulated multi-focus images, which are different from real-world multi-focus images, thus the generalization abilities of these algorithms are not good.

The results on Lytro dataset indicate that various MFIF algorithms may have very different performances on different evaluation metrics, therefore it is necessary to use different kinds of metrics when evaluating MFIF approaches.~Besides, although the qualitative performances are not very consistent with the overall quantitative performances, they are consistent with the human perception inspired metrics to some extent, especially $Q_{CB}$ and $Q_{CV}$.

Xu et al.~\cite{xu2020mffw} pointed out that the defocus spread effect (DSE) is not obvious on the images of the Lytro dataset, thus the fused images produced by many algorithms have no significant visual differences.~To further compare MFIF algorithms, the comparison of fusion results on the whole MFIFB dataset will be presented in the following Section. 

\subsection{Results on the whole MFIFB dataset}
\label{subsec:whole}

\subsubsection{Qualitative performance comparison}
\label{subsec:qualitative}
Figure \ref{fig:qualitative-mmfw12} presents the qualitative (visual) performance comparison of 30 fusion methods on the \textit{MMFW12} image pair.~One can see that this case is more difficult than the \textit{Lytro19} case, since many algorithms do not produced satisfactory fused image on this image pair.~To be more specific, some methods, including BGSC, CBF, CSR, DCT\_Corr, DCT\_EOL, DPRL, DSIFT, DWTDE, ECNN, NSCT\_SR, QB, SESF, SVDDCT, and TF, have obvious visual artifacts.~Besides, some algorithms show obvious color distortion, namely MGFF, MST\_SR, PR\_SR.~Because the rest of algorithms do not show obvious visual artifacts and color distortion, thus two focused/defocused boundary areas are illustrated in the magnified plots to see more details.~As can be seen, BFMF, GFDF, IFM, MFM, PCANet do not handle the boundary area contained in the red box well.~Besides, BFMF, CNN, GD, GFDF, IFCNN, IFM, MFM, MSVD, PCANet and QB cannot deal with the DSE in the boundary area contained in the green box as can be seen from the magnified plots.~Overall, ASR, GFF and MWGF show good performance on the \textit{MMFW12} image pair.

\begin{table*}
	\begin{center}
		\caption{Running time of algorithms in MFIFB (seconds per image pair)}
		\label{table:runtime}
		\footnotesize
		\begin{tabular}{llllll}		
			\hline
			Method      & Average running time     & Category & Method      & Average running time   & Category    \\ \hline
			ASR \cite{liu2014simultaneous}    &  549.95  &  SR-based    & CBF \cite{kumar2015image}   & 21.11  &         Multi-scale-based	\\	
			CSR \cite{liu2016image}     &    466.27  &   SR-based & DCT\textunderscore Corr \cite{amin2018multi} & 0.34 &   Multi-scale-based     \\ 
			CNN \cite{liu2017multi}  &184.78 &     DL-based & DCT\textunderscore EOL \cite{amin2018multi} &    0.24   &Multi-scale-based \\
			DRPL \cite{li2020drpl}   & 0.17 & DL-based & 	DWTDE \cite{liu2013multi}  &7.84 &Multi-scale-based 			 \\
			ECNN \cite{amin-naji2019ensemble} & 62.92  &DL-based & GD \cite{paul2016multi} & 0.55  &  Multi-scale-based   		       \\
			IFCNN \cite{zhang2020ifcnn}  &0.03 &  DL-based  &	MSVD \cite{naidu2011image}   &0.92 &Multi-scale-based\\	
			PCANet \cite{song2018multi}  & 20.77  & DL-based 	&MWGF \cite{zhou2014multi} &  2.76 &Multi-scale-based      \\
			SESF \cite{ma2019sesf}  &    0.16   &   DL-based    &			SVDDCT \cite{amin2017multi} &1.09 &  Multi-scale-based\\
			BGSC \cite{tian2011multi}         &  6.52    &  Spatial domain-based &GFDF \cite{qiu2019guided}   &  0.23   &   Edge-preserving-based   \\
			BFMF \cite{zhang2017boundary}     &  1.36    &  Spatial domain-based & GFF \cite{li2013image}  & 0.42   &   Edge-preserving-based        \\
			
			DSIFT \cite{liu2015multi} & 7.53    &Spatial domain-based  & 	MFM \cite{ma2017multi} & 1.45 & Edge-preserving-based \\
			IFM \cite{li2013imageb}   & 2.18 &  Spatial domain-based  & 	MGFF \cite{bavirisetti2019multi}   &   1.17   & Edge-preserving-based  \\
			QB \cite{bai2015quadtree} &1.07 & Spatial domain-based  & MST\textunderscore SR \cite{liu2015general}  &  0.75  & Hybrid  \\
			TF \cite{ma2019multi}  &0.48 &  Spatial domain-based 	&		NSCT\textunderscore SR \cite{liu2015general}   & 91.95  & Hybrid  \\	
			SFMD \cite{li2015multi}   & 0.81  & Subspace-based 	&		RP\textunderscore SR \cite{liu2015general}   & 0.81 & Hybrid  \\		
			
			\hline
		\end{tabular}
	\end{center}
\end{table*}

\subsubsection{Quantitative performance comparison}
\label{subsec:quantitative}
Table \ref{table:metrics_average} presents the average value of 20 evaluation metrics for all methods on the whole MFIFB dataset.~From the table one can see that the top three methods on the whole MFIFB dataset are QB, SFMD and GFDF, respectively.~Although SFMD and QB have the same number of top three metric values, QB is ranked the first while SFMD the second.~This is because that QB performs well in information theory-based,  image structural similarity-based and human perception inspired metrics.~In contrast, SFMD only shows good performances in image feature-based metrics but performs poorly in other kinds of metrics.

The performances of MFIF algorithms on the whole MFIF dataset are very different from that on the Lytro subset.~First, the spatial domain-based approaches perform better than the transform domain-based ones.~Second, deep learning-based methods have worse performances on the whole MFIFB dataset than on the Lytro dataset.~Specifically, the best deep learning-based methods, namely DPRL, only ranks the fifth on the whole dataset.~Apart from Lytro, the MFIFB dataset also contains other subsets such as MFFW and those proposed by Savic et al.~\cite{savic2012multifocus} and Aymaz et al.~\cite{aymaz2020multi}.~In other words, the whole MFIFB dataset is more challenging than the Lytro dataset.~The reason why the performances of deep learning-based approaches degrade is that they do not perform well on the remaining subsets in MFIFB except Lytro.~For instance, the MFFW dataset has strong defocus spread effect but the simulated training data of deep learning-based methods do not have, so they cannot learn how to handle defocus spread effect.

\subsection{Running time comparison}
\label{subsec:runtime}
Table \ref{table:runtime} lists the average running time of all algorithms integrated in MFIFB.~As can be seen, the running time of image fusion methods varies significantly from one to another.~Generally speaking, SR-based methods are most computational expensive, which take more than 7 minutes to fuse an image pair.~Besides, transform domain-based methods are generally faster than their spatial domain-based counterparts except some cases like CBF.~Regarding the deep learning-based algorithms, the computational efficiency also varies greatly.~For example, the running time of CNN is more than 6000 times that of IFCNN.~Note that all deep learning-based algorithms in MFIFB do not update the model online.

\section{Concluding remarks}
\label{sec:conclusion}
In this paper, a multi-focus image fusion benchmark (MFIFB), which includes a dataset of 105 image pairs, a code library of 30 algorithms, 20 evaluation metrics and all results is proposed.~To the best of our knowledge, this is the first multi-focus image fusion benchmark to date.~This benchmark facilitates better understanding of the state-of-the-art MFIF approaches and provides a platform for comparing performance among algorithms fairly and comprehensively.~It should be noted that, the proposed MFIFB can be easily extended to contain more images, fusion algorithms and more evaluation metrics.

In the literature, MFIF algorithms are usually tested on a small number of images and compared with a very limited number of algorithms using just several evaluation metrics, therefore the performance comparison may not be fair and comprehensive.~This makes it difficult to understand the state-of-the-art of the MFIF field and hinders the future development of new algorithms.~To solve this issue, in this study extensive experiments have been carried out based on MFIFB to evaluate the performance of all integrated fusion algorithms.

We have several observations on the status of MFIF based on the experimental results.~First, unlike other fields in computer vision where deep learning is almost the dominant method, deep learning methods do not provide very competitive performances on challenging MFIF datasets at the moment.~Conventional methods, namely spatial domain-based and transform domain-based ones, still have good performances.~This is very supervising because many deep learning-based MFIF methods were claimed to have the state-of-the-art performances.~However, this is not really true on challenging MFIF dataset according to our experiments using the proposed MFIFB.~The possible reason is that most deep learning-based MFIF algorithms were trained on simulated MFIF data which do not show much defocus spread effect, thus these algorithms cannot generalize well to other real-world MFIF dataset.~Besides, those methods were only compared with a small number of methods using several evaluation metrics on a small dataset which does not have much defocus spread effect, thus the performances were not fully evaluated.~However, due to the strong representation ability and  end-to-end property of deep learning, we believe that the deep learning-based approach will be an important research direction in future.~Second, a MFIF algorithm usually cannot have good performances in all aspects in terms of evaluation metrics.~Some algorithms may achieve good values in information theory-based metrics while others may perform well in other kinds of metrics.~Therefore, it is very important to use several kinds of evaluation metrics when conducting quantitative performance comparison for MFIF algorithms.~Finally, the results of qualitative and quantitative comparisons may not be consistent for a MFIF algorithm, therefore they are both crucial when evaluating a MFIF method.

We will continue extending MFIFB by including more image pairs, algorithms and metrics.~We hope that MFIFB can serve as a good starting point for researchers who are interested in multi-focus image fusion.

\section*{Acknowledgment}	
The author would like to thank Mr Shuang Xu from Xi'an Jiao Tong University for providing the MFFW dataset.
	
	\ifCLASSOPTIONcaptionsoff
	\newpage
	\fi

	\bibliographystyle{IEEEtran}
	\bibliography{../../../xingchen}

\begin{thebibliography}{10}
\providecommand{\url}[1]{#1}
\csname url@samestyle\endcsname
\providecommand{\newblock}{\relax}
\providecommand{\bibinfo}[2]{#2}
\providecommand{\BIBentrySTDinterwordspacing}{\spaceskip=0pt\relax}
\providecommand{\BIBentryALTinterwordstretchfactor}{4}
\providecommand{\BIBentryALTinterwordspacing}{\spaceskip=\fontdimen2\font plus
\BIBentryALTinterwordstretchfactor\fontdimen3\font minus
  \fontdimen4\font\relax}
\providecommand{\BIBforeignlanguage}[2]{{%
\expandafter\ifx\csname l@#1\endcsname\relax
\typeout{** WARNING: IEEEtran.bst: No hyphenation pattern has been}%
\typeout{** loaded for the language `#1'. Using the pattern for}%
\typeout{** the default language instead.}%
\else
\language=\csname l@#1\endcsname
\fi
#2}}
\providecommand{\BIBdecl}{\relax}
\BIBdecl

\bibitem{liu2015multi}
Y.~Liu, S.~Liu, and Z.~Wang, ``Multi-focus image fusion with dense sift,''
  \emph{Information Fusion}, vol.~23, pp. 139--155, 2015.

\bibitem{de2013multi}
I.~De and B.~Chanda, ``Multi-focus image fusion using a morphology-based focus
  measure in a quad-tree structure,'' \emph{Information Fusion}, vol.~14,
  no.~2, pp. 136--146, 2013.

\bibitem{li2006region}
M.~Li, W.~Cai, and Z.~Tan, ``A region-based multi-sensor image fusion scheme
  using pulse-coupled neural network,'' \emph{Pattern Recognition Letters},
  vol.~27, no.~16, pp. 1948--1956, 2006.

\bibitem{zhang2018robust}
Q.~Zhang, T.~Shi, F.~Wang, R.~S. Blum, and J.~Han, ``Robust sparse
  representation based multi-focus image fusion with dictionary construction
  and local spatial consistency,'' \emph{Pattern Recognition}, vol.~83, pp.
  299--313, 2018.

\bibitem{zhang2018sparse}
Q.~Zhang, Y.~Liu, R.~S. Blum, J.~Han, and D.~Tao, ``Sparse representation based
  multi-sensor image fusion for multi-focus and multi-modality images: A
  review,'' \emph{Information Fusion}, vol.~40, pp. 57--75, 2018.

\bibitem{amin2018multi}
M.~Amin-Naji and A.~Aghagolzadeh, ``Multi-focus image fusion in dct domain
  using variance and energy of laplacian and correlation coefficient for visual
  sensor networks,'' \emph{Journal of AI and Data Mining}, vol.~6, no.~2, pp.
  233--250, 2018.

\bibitem{kou2018multi}
L.~Kou, L.~Zhang, K.~Zhang, J.~Sun, Q.~Han, and Z.~Jin, ``A multi-focus image
  fusion method via region mosaicking on laplacian pyramids,'' \emph{PloS one},
  vol.~13, no.~5, 2018.

\bibitem{kumar2015image}
B.~K. Shreyamsha~Kumar, ``Image fusion based on pixel significance using cross
  bilateral filter,'' \emph{Signal, Image and Video Processing}, vol.~9, no.~5,
  pp. 1193--1204, Jul 2015.

\bibitem{zhai2020multi}
H.~Zhai and Y.~Zhuang, ``Multi-focus image fusion method using energy of
  laplacian and a deep neural network,'' \emph{Applied Optics}, vol.~59, no.~6,
  pp. 1684--1694, 2020.

\bibitem{zhang2020ifcnn}
Y.~Zhang, Y.~Liu, P.~Sun, H.~Yan, X.~Zhao, and L.~Zhang, ``{IFCNN: A general
  image fusion framework based on convolutional neural network},''
  \emph{Information Fusion}, vol.~54, no. August 2018, pp. 99--118, 2020.

\bibitem{wang2020novel}
C.~Wang, Z.~Zhao, Q.~Ren, Y.~Xu, and Y.~Yu, ``A novel multi-focus image fusion
  by combining simplified very deep convolutional networks and patch-based
  sequential reconstruction strategy,'' \emph{Applied Soft Computing}, p.
  106253, 2020.

\bibitem{xu2020deep}
H.~Xu, F.~Fan, H.~Zhang, Z.~Le, and J.~Huang, ``A deep model for multi-focus
  image fusion based on gradients and connected regions,'' \emph{IEEE Access},
  vol.~8, pp. 26\,316--26\,327, 2020.

\bibitem{yan2018unsupervised}
X.~Yan, S.~Z. Gilani, H.~Qin, and A.~Mian, ``Unsupervised deep multi-focus
  image fusion,'' \emph{arXiv preprint arXiv:1806.07272}, 2018.

\bibitem{ma2019sesf}
B.~Ma, X.~Ban, H.~Huang, and Y.~Zhu, ``Sesf-fuse: An unsupervised deep model
  for multi-focus image fusion,'' \emph{arXiv preprint arXiv:1908.01703}, 2019.

\bibitem{mustafa2019dense}
H.~T. Mustafa, F.~Liu, J.~Yang, Z.~Khan, and Q.~Huang, ``Dense multi-focus
  fusion net: A deep unsupervised convolutional network for multi-focus image
  fusion,'' in \emph{International Conference on Artificial Intelligence and
  Soft Computing}.\hskip 1em plus 0.5em minus 0.4em\relax Springer, 2019, pp.
  153--163.

\bibitem{jung2020unsupervised}
H.~Jung, Y.~Kim, H.~Jang, N.~Ha, and K.~Sohn, ``{Unsupervised Deep Image Fusion
  with Structure Tensor Representations},'' \emph{IEEE Transactions on Image
  Processing}, vol.~29, pp. 3845--3858, 2020.

\bibitem{li2018convolutional}
H.~Li, R.~Nie, D.~Zhou, and X.~Gou, ``Convolutional neural network based
  multi-focus image fusion,'' in \emph{Proceedings of the 2018 2nd
  International Conference on Algorithms, Computing and Systems}, 2018, pp.
  148--154.

\bibitem{du2017image}
C.~Du and S.~Gao, ``Image segmentation-based multi-focus image fusion through
  multi-scale convolutional neural network,'' \emph{IEEE access}, vol.~5, pp.
  15\,750--15\,761, 2017.

\bibitem{guo2019fusegan}
X.~Guo, R.~Nie, J.~Cao, D.~Zhou, L.~Mei, K.~He, S.~Member, R.~Nie, J.~Cao, and
  D.~Zhou, ``{FuseGAN: Learning to fuse Multi-focus Image via Conditional
  Generative Adversarial Network},'' \emph{IEEE Transactions on Multimedia},
  vol.~21, no.~8, pp. 1--1, 2019.

\bibitem{amin-naji2019ensemble}
M.~Amin-Naji, A.~Aghagolzadeh, and M.~Ezoji, ``{Ensemble of CNN for Multi-Focus
  Image Fusion},'' \emph{Information Fusion}, vol.~51, no. February, pp.
  201--214, 2019.

\bibitem{li2013image}
S.~Li, X.~Kang, and J.~Hu, ``Image fusion with guided filtering,'' \emph{IEEE
  Transactions on Image processing}, vol.~22, no.~7, pp. 2864--2875, 2013.

\bibitem{liu2015general}
Y.~Liu, S.~Liu, and Z.~Wang, ``A general framework for image fusion based on
  multi-scale transform and sparse representation,'' \emph{Information Fusion},
  vol.~24, pp. 147--164, 2015.

\bibitem{bai2015quadtree}
X.~Bai, Y.~Zhang, F.~Zhou, and B.~Xue, ``Quadtree-based multi-focus image
  fusion using a weighted focus-measure,'' \emph{Information Fusion}, vol.~22,
  pp. 105--118, 2015.

\bibitem{zhang2016robustmulti}
Q.~Zhang and M.~D. Levine, ``Robust multi-focus image fusion using multi-task
  sparse representation and spatial context,'' \emph{IEEE Transactions on Image
  Processing}, vol.~25, no.~5, pp. 2045--2058, 2016.

\bibitem{liu2017multi}
Y.~Liu, X.~Chen, H.~Peng, and Z.~Wang, ``Multi-focus image fusion with a deep
  convolutional neural network,'' \emph{Information Fusion}, vol.~36, pp.
  191--207, 2017.

\bibitem{zhang2017boundary}
Y.~Zhang, X.~Bai, and T.~Wang, ``Boundary finding based multi-focus image
  fusion through multi-scale morphological focus-measure,'' \emph{Information
  fusion}, vol.~35, pp. 81--101, 2017.

\bibitem{tang2018pixel}
H.~Tang, B.~Xiao, W.~Li, and G.~Wang, ``Pixel convolutional neural network for
  multi-focus image fusion,'' \emph{Information Sciences}, vol. 433, pp.
  125--141, 2018.

\bibitem{farid2019multi}
M.~S. Farid, A.~Mahmood, and S.~A. Al-Maadeed, ``Multi-focus image fusion using
  content adaptive blurring,'' \emph{Information fusion}, vol.~45, pp. 96--112,
  2019.

\bibitem{bouzos2019conditional}
O.~Bouzos, I.~Andreadis, and N.~Mitianoudis, ``Conditional random field model
  for robust multi-focus image fusion,'' \emph{IEEE Transactions on Image
  Processing}, vol.~28, no.~11, pp. 5636--5648, 2019.

\bibitem{li2020drpl}
J.~Li, X.~Guo, G.~Lu, B.~Zhang, Y.~Xu, F.~Wu, and D.~Zhang, ``Drpl: Deep
  regression pair learning for multi-focus image fusion,'' \emph{IEEE
  Transactions on Image Processing}, vol.~29, pp. 4816--4831, 2020.

\bibitem{xu2020fusiondn}
H.~Xu, J.~Ma, Z.~Le, J.~Jiang, and X.~Guo, ``Fusiondn: A unified densely
  connected network for image fusion,'' in \emph{Thirty-Fourth AAAI Conference
  on Artificial Intelligence}, 2020.

\bibitem{zhang2020PMGI}
H.~Zhang, H.~Xu, Y.~Xiao, X.~Guo, and J.~Ma, ``Rethinking the image fusion: A
  fast unified image fusion network based on proportional maintenance of
  gradient and intensity,'' in \emph{Proceedings of the AAAI Conference on
  Artificial Intelligence}, 2020.

\bibitem{nejati2015multi}
M.~Nejati, S.~Samavi, and S.~Shirani, ``Multi-focus image fusion using
  dictionary-based sparse representation,'' \emph{Information Fusion}, vol.~25,
  pp. 72--84, 2015.

\bibitem{wu2015object}
Y.~Wu, J.~Lim, and M.-H. Yang, ``Object tracking benchmark,'' \emph{IEEE
  Transactions on Pattern Analysis and Machine Intelligence}, vol.~37, no.~9,
  pp. 1834--1848, 2015.

\bibitem{VOT_TPAMI}
M.~Kristan, J.~Matas, A.~Leonardis, T.~Vojir, R.~Pflugfelder, G.~Fernandez,
  G.~Nebehay, F.~Porikli, and L.~\v{C}ehovin, ``A novel performance evaluation
  methodology for single-target trackers,'' \emph{IEEE Transactions on Pattern
  Analysis and Machine Intelligence}, vol.~38, no.~11, pp. 2137--2155, Nov
  2016.

\bibitem{xu2020mffw}
S.~Xu, X.~Wei, C.~Zhang, J.~Liu, and J.~Zhang, ``Mffw: A new dataset for
  multi-focus image fusion,'' \emph{arXiv preprint arXiv:2002.04780}, 2020.

\bibitem{deng2019deep}
X.~Deng and P.~L. Dragotti, ``{Deep Convolutional Neural Network for
  Multi-modal Image Restoration and Fusion},'' pp. 1--15, 2019.

\bibitem{fu2020novel}
G.-P. Fu, S.-H. Hong, F.-L. Li, and L.~Wang, ``A novel multi-focus image fusion
  method based on distributed compressed sensing,'' \emph{Journal of Visual
  Communication and Image Representation}, vol.~67, p. 102760, 2020.

\bibitem{yang2010multi}
B.~{Yang} and S.~{Li}, ``Multifocus image fusion and restoration with sparse
  representation,'' \emph{IEEE Transactions on Instrumentation and
  Measurement}, vol.~59, no.~4, pp. 884--892, 2010.

\bibitem{liu2014simultaneous}
Y.~Liu and Z.~Wang, ``Simultaneous image fusion and denoising with adaptive
  sparse representation,'' \emph{IET Image Processing}, vol.~9, no.~5, pp.
  347--357, 2014.

\bibitem{wang2003multifocus}
W.-W. Wang, P.-L. Shui, and G.-X. Song, ``Multifocus image fusion in wavelet
  domain,'' in \emph{Proceedings of the 2003 International Conference on
  Machine Learning and Cybernetics (IEEE Cat. No. 03EX693)}, vol.~5.\hskip 1em
  plus 0.5em minus 0.4em\relax IEEE, 2003, pp. 2887--2890.

\bibitem{zhi2004wavelet}
J.~Zhi-guo, H.~Dong-bing, C.~Jin, and Z.~Xiao-kuan, ``A wavelet based algorithm
  for multi-focus micro-image fusion,'' in \emph{Third International Conference
  on Image and Graphics (ICIG'04)}.\hskip 1em plus 0.5em minus 0.4em\relax
  IEEE, 2004, pp. 176--179.

\bibitem{bavirisetti2017multi}
D.~P. Bavirisetti, G.~Xiao, and G.~Liu, ``Multi-sensor image fusion based on
  fourth order partial differential equations,'' in \emph{2017 20th
  International Conference on Information Fusion (Fusion)}.\hskip 1em plus
  0.5em minus 0.4em\relax IEEE, 2017, pp. 1--9.

\bibitem{chen2017robust}
Y.~Chen, J.~Guan, and W.-K. Cham, ``Robust multi-focus image fusion using edge
  model and multi-matting,'' \emph{IEEE Transactions on Image Processing},
  vol.~27, no.~3, pp. 1526--1541, 2017.

\bibitem{zhao2017multisensor}
W.~Zhao, H.~Lu, and D.~Wang, ``Multisensor image fusion and enhancement in
  spectral total variation domain,'' \emph{IEEE Transactions on Multimedia},
  vol.~20, no.~4, pp. 866--879, 2017.

\bibitem{zhao2018multi}
W.~Zhao, D.~Wang, and H.~Lu, ``Multi-focus image fusion with a natural
  enhancement via a joint multi-level deeply supervised convolutional neural
  network,'' \emph{IEEE Transactions on Circuits and Systems for Video
  Technology}, vol.~29, no.~4, pp. 1102--1115, 2018.

\bibitem{yang2019multilevel}
Y.~Yang, Z.~Nie, S.~Huang, P.~Lin, and J.~Wu, ``Multilevel features
  convolutional neural network for multifocus image fusion,'' \emph{IEEE
  Transactions on Computational Imaging}, vol.~5, no.~2, pp. 262--273, 2019.

\bibitem{deshmukh2017multi}
V.~Deshmukh, A.~Khaparde, and S.~Shaikh, ``Multi-focus image fusion using deep
  belief network,'' in \emph{International Conference on Information and
  Communication Technology for Intelligent Systems}.\hskip 1em plus 0.5em minus
  0.4em\relax Springer, 2017, pp. 233--241.

\bibitem{savic2012multifocus}
S.~Savi{\'c} and Z.~Babi{\'c}, ``Multifocus image fusion based on empirical
  mode decomposition,'' in \emph{19th IEEE International Conference on Systems,
  Signals and Image Processing (IWSSIP)}, 2012.

\bibitem{aymaz2020multi}
S.~Aymaz, C.~K{\"o}se, and {\c{S}}.~Aymaz, ``Multi-focus image fusion for
  different datasets with super-resolution using gradient-based new fusion
  rule,'' \emph{Multimedia Tools and Applications}, pp. 1--40, 2020.

\bibitem{tian2011multi}
J.~Tian, L.~Chen, L.~Ma, and W.~Yu, ``Multi-focus image fusion using a
  bilateral gradient-based sharpness criterion,'' \emph{Optics communications},
  vol. 284, no.~1, pp. 80--87, 2011.

\bibitem{liu2016image}
Y.~Liu, X.~Chen, R.~K. Ward, and Z.~J. Wang, ``Image fusion with convolutional
  sparse representation,'' \emph{IEEE signal processing letters}, vol.~23,
  no.~12, pp. 1882--1886, 2016.

\bibitem{liu2013multi}
Y.~Liu and Z.~Wang, ``Multi-focus image fusion based on wavelet transform and
  adaptive block,'' \emph{Journal of image and graphics}, vol.~18, no.~11, pp.
  1435--1444, 2013.

\bibitem{paul2016multi}
S.~Paul, I.~S. Sevcenco, and P.~Agathoklis, ``Multi-exposure and multi-focus
  image fusion in gradient domain,'' \emph{Journal of Circuits, Systems and
  Computers}, vol.~25, no.~10, p. 1650123, 2016.

\bibitem{qiu2019guided}
X.~Qiu, M.~Li, L.~Zhang, and X.~Yuan, ``Guided filter-based multi-focus image
  fusion through focus region detection,'' \emph{Signal Processing: Image
  Communication}, vol.~72, pp. 35--46, 2019.

\bibitem{li2013imageb}
S.~Li, X.~Kang, J.~Hu, and B.~Yang, ``Image matting for fusion of multi-focus
  images in dynamic scenes,'' \emph{Information Fusion}, vol.~14, no.~2, pp.
  147--162, 2013.

\bibitem{ma2017multi}
J.~Ma, Z.~Zhou, B.~Wang, and M.~Dong, ``Multi-focus image fusion based on
  multi-scale focus measures and generalized random walk,'' in \emph{2017 36th
  Chinese Control Conference (CCC)}.\hskip 1em plus 0.5em minus 0.4em\relax
  IEEE, 2017, pp. 5464--5468.

\bibitem{bavirisetti2019multi}
D.~P. Bavirisetti, G.~Xiao, J.~Zhao, R.~Dhuli, and G.~Liu, ``Multi-scale guided
  image and video fusion: A fast and efficient approach,'' \emph{Circuits,
  Systems, and Signal Processing}, vol.~38, no.~12, pp. 5576--5605, Dec 2019.

\bibitem{naidu2011image}
V.~Naidu, ``Image fusion technique using multi-resolution singular value
  decomposition,'' \emph{Defence Science Journal}, vol.~61, no.~5, pp.
  479--484, 2011.

\bibitem{zhou2014multi}
Z.~Zhou, S.~Li, and B.~Wang, ``Multi-scale weighted gradient-based fusion for
  multi-focus images,'' \emph{Information Fusion}, vol.~20, pp. 60--72, 2014.

\bibitem{song2018multi}
X.~Song and X.-J. Wu, ``{Multi-focus Image Fusion with PCA Filters of
  PCANet},'' in \emph{IAPR Workshop on Multimodal Pattern Recognition of Social
  Signals in Human-Computer Interaction}.\hskip 1em plus 0.5em minus
  0.4em\relax Springer, 2018, pp. 1--17.

\bibitem{li2015multi}
H.~Li, L.~Li, and J.~Zhang, ``Multi-focus image fusion based on sparse feature
  matrix decomposition and morphological filtering,'' \emph{Optics
  Communications}, vol. 342, pp. 1--11, 2015.

\bibitem{amin2017multi}
M.~Amin-Naji, P.~Ranjbar-Noiey, and A.~Aghagolzadeh, ``Multi-focus image fusion
  using singular value decomposition in dct domain,'' in \emph{2017 10th
  Iranian Conference on Machine Vision and Image Processing (MVIP)}.\hskip 1em
  plus 0.5em minus 0.4em\relax IEEE, 2017, pp. 45--51.

\bibitem{ma2019multi}
J.~Ma, Z.~Zhou, B.~Wang, L.~Miao, and H.~Zong, ``Multi-focus image fusion using
  boosted random walks-based algorithm with two-scale focus maps,''
  \emph{Neurocomputing}, vol. 335, pp. 9--20, 2019.

\bibitem{liu2012objective}
Z.~Liu, E.~Blasch, Z.~Xue, J.~Zhao, R.~Laganiere, and W.~Wu, ``Objective
  assessment of multiresolution image fusion algorithms for context enhancement
  in night vision: A comparative study,'' \emph{IEEE Transactions on Pattern
  Analysis and Machine Intelligence}, vol.~34, pp. 94--109, 2012.

\bibitem{bulanon2009image}
D.~M. Bulanon, T.~Burks, and V.~Alchanatis, ``Image fusion of visible and
  thermal images for fruit detection,'' \emph{Biosystems Engineering}, vol.
  103, no.~1, pp. 12--22, 2009.

\bibitem{aardt2008assessment}
V.~Aardt and Jan, ``Assessment of image fusion procedures using entropy, image
  quality, and multispectral classification,'' \emph{Journal of Applied Remote
  Sensing}, vol.~2, no.~1, p. 023522, 2008.

\bibitem{hossny2008comments}
M.~Hossny, S.~Nahavandi, and D.~Creighton, ``Comments on'information measure
  for performance of image fusion','' \emph{Electronics letters}, vol.~44,
  no.~18, pp. 1066--1067, 2008.

\bibitem{jagalingam2015review}
P.~Jagalingam and A.~V. Hegde, ``A review of quality metrics for fused image,''
  \emph{Aquatic Procedia}, vol.~4, no. Icwrcoe, pp. 133--142, 2015.

\bibitem{wang2005nonlinear}
Q.~Wang, Y.~Shen, and J.~Q. Zhang, ``A nonlinear correlation measure for
  multivariable data set,'' \emph{Physica D: Nonlinear Phenomena}, vol. 200,
  no. 3-4, pp. 287--295, 2005.

\bibitem{wang2008performance}
Q.~Wang, Y.~Shen, and J.~Jin, ``Performance evaluation of image fusion
  techniques,'' \emph{Image fusion: algorithms and applications}, vol.~19, pp.
  469--492, 2008.

\bibitem{cvejic2006image}
N.~Cvejic, C.~Canagarajah, and D.~Bull, ``Image fusion metric based on mutual
  information and tsallis entropy,'' \emph{Electronics letters}, vol.~42,
  no.~11, pp. 626--627, 2006.

\bibitem{cui2015detail}
G.~Cui, H.~Feng, Z.~Xu, Q.~Li, and Y.~Chen, ``Detail preserved fusion of
  visible and infrared images using regional saliency extraction and
  multi-scale image decomposition,'' \emph{Optics Communications}, vol. 341,
  pp. 199 -- 209, 2015.

\bibitem{rajalingam2018hybrid}
B.~Rajalingam and R.~Priya, ``Hybrid multimodality medical image fusion
  technique for feature enhancement in medical diagnosis,'' \emph{International
  Journal of Engineering Science Invention}, 2018.

\bibitem{xydeas2000objective}
C.~S. Xydeas and P.~V. V., ``Objective image fusion performance measure,''
  \emph{Military Technical Courier}, vol.~36, no.~4, pp. 308--309, 2000.

\bibitem{zhao2007performance}
J.~Zhao, R.~Laganiere, and Z.~Liu, ``Performance assessment of combinative
  pixel-level image fusion based on an absolute feature measurement,''
  \emph{International Journal of Innovative Computing, Information and
  Control}, vol.~3, no.~6, pp. 1433--1447, 2007.

\bibitem{rao1997fibre}
Y.-J. Rao, ``In-fibre bragg grating sensors,'' \emph{Measurement science and
  technology}, vol.~8, no.~4, p. 355, 1997.

\bibitem{eskicioglu1995image}
A.~M. Eskicioglu and P.~S. Fisher, ``Image quality measures and their
  performance,'' \emph{IEEE Transactions on communications}, vol.~43, no.~12,
  pp. 2959--2965, 1995.

\bibitem{cvejic2005similarity}
N.~Cvejic, A.~Loza, D.~Bull, and N.~Canagarajah, ``A similarity metric for
  assessment of image fusion algorithms,'' \emph{International journal of
  signal processing}, vol.~2, no.~3, pp. 178--182, 2005.

\bibitem{piella2003new}
G.~Piella and H.~Heijmans, ``A new quality metric for image fusion,'' in
  \emph{Proceedings 2003 International Conference on Image Processing (Cat. No.
  03CH37429)}, vol.~3.\hskip 1em plus 0.5em minus 0.4em\relax IEEE, 2003, pp.
  III--173.

\bibitem{yang2008novel}
C.~Yang, J.-Q. Zhang, X.-R. Wang, and X.~Liu, ``A novel similarity based
  quality metric for image fusion,'' \emph{Information Fusion}, vol.~9, no.~2,
  pp. 156--160, 2008.

\bibitem{wang2004image}
Z.~Wang, A.~C. Bovik, H.~R. Sheikh, E.~P. Simoncelli \emph{et~al.}, ``Image
  quality assessment: from error visibility to structural similarity,''
  \emph{IEEE transactions on image processing}, vol.~13, no.~4, pp. 600--612,
  2004.

\bibitem{chen2009new}
Y.~Chen and R.~S. Blum, ``A new automated quality assessment algorithm for
  image fusion,'' \emph{Image and vision computing}, vol.~27, no.~10, pp.
  1421--1432, 2009.

\bibitem{chen2007human}
H.~Chen and P.~K. Varshney, ``A human perception inspired quality metric for
  image fusion based on regional information,'' \emph{Information fusion},
  vol.~8, no.~2, pp. 193--207, 2007.

\bibitem{han2013new}
Y.~Han, Y.~Cai, Y.~Cao, and X.~Xu, ``{A new image fusion performance metric
  based on visual information fidelity},'' \emph{Information Fusion}, vol.~14,
  no.~2, pp. 127--135, 2013.

\bibitem{ma2019infrared}
J.~Ma, Y.~Ma, and C.~Li, ``Infrared and visible image fusion methods and
  applications: A survey,'' \emph{Information Fusion}, vol.~45, pp. 153--178,
  2019.

\end{thebibliography}

\end{document}